\documentclass[9pt,shortpaper,twoside,web]{ieeecolor2}
\usepackage{generic}
\usepackage{cite}
\usepackage{amsmath,amssymb,amsfonts}
\usepackage{algorithmic}
\usepackage{graphicx}
\usepackage{textcomp}
\usepackage{multirow}
\usepackage{booktabs}
\usepackage[dvipsnames]{xcolor}
\let\labelindent\relax
\usepackage{enumitem}
\usepackage{url}
\usepackage{listings}
\def\BibTeX{{\rm B\kern-.05em{\sc i\kern-.025em b}\kern-.08em
    T\kern-.1667em\lower.7ex\hbox{E}\kern-.125emX}}
\usepackage[hidelinks]{hyperref}
\usepackage{cleveref}

\newcommand{\algname}{InterGenEHR$^+$}
\newcommand{\algnameAbbrv}{{\bf IGE}}
\def \citep{\cite}

\definecolor{revisionblue}{RGB}{0,76,153}
\newcommand{\revisiontwo}[1]{#1}
\newcommand{\revision}[1]{#1}

\begin{document}
\title{InterGenEHR$^+$: An Interactive Framework for Natural Language-Conditioned Generation of Longitudinal Patient Trajectories
}

\author{Yu Yang*, Jiafeng Song*, Baraa F Abed, Christina Creel, Sagar B Dave, Henry Foote, Lisa Lima, Victor Moas, Craig S Jabaley, Rishikesan Kamaleswaran, and Pan Xu
\thanks{*Yu Yang and Jiafeng Song contributed equally to this work.}
\thanks{Yu Yang is with the Department of Electrical and Computer Engineering, Duke University, Durham, NC (e-mail: yu.yang@duke.edu).}
\thanks{Jiafeng Song is with the Department of Biomedical Engineering, Duke University, Durham, NC (e-mail: stuart.song@duke.edu).}
\thanks{Baraa F Abed is with the Department of Electrical and Computer Engineering, Duke University, Durham, NC (e-mail: baraa.abed@duke.edu).}
\thanks{Christina Creel is with the Department of Emergency Medicine, Emory University, Atlanta, GA (e-mail: maria.christina.creel-bulos@emory.edu).}
\thanks{Sagar B Dave is with the Department of Emergency Medicine, Emory University, Atlanta, GA (e-mail: sagar.bharat.dave@emory.edu).}
\thanks{Henry Foote is with the Department of Pediatrics, Duke University, Durham, NC (e-mail: henry.foote@duke.edu).}
\thanks{Lisa Lima is with the Department of Emergency Medicine, Emory University, Atlanta, GA (e-mail: lisa.lima@emory.edu).}
\thanks{Victor Moas is with the Department of Surgery, Duke University, Durham, NC (e-mail: victor.moas@duke.edu).}
\thanks{Craig S Jabaley is with the Department of Anesthesiology, Emory University, Atlanta, GA (e-mail: csjabaley@emory.edu).}
\thanks{Rishikesan Kamaleswaran is with the Department of Surgery, Duke University, Durham, NC (e-mail: r.kamaleswaran@duke.edu).}
\thanks{Pan Xu is with the Department of Biostatistics and Bioinformatics, Duke University, Durham, NC (e-mail: pan.xu@duke.edu).}}

\IEEEaftertitletext{%
\vspace{-3\baselineskip}%
\begin{minipage}{\textwidth}
\normalfont\scriptsize\raggedright
\textbf{Author-accepted manuscript.} Published as: Y. Yang et al.,
``InterGenEHR$^+$: An Interactive Framework for Natural Language-Conditioned
Generation of Longitudinal Patient Trajectories,''
\textit{IEEE Transactions on Biomedical Engineering}, early access,
Nov. 19, 2026, doi: \url{https://doi.org/10.1109/TBME.2026.3753060}.
\\[0.5ex]
\textcopyright{} 2026 IEEE. Personal use of this material is permitted.
Permission from IEEE must be obtained for all other uses, in any current or
future media, including reprinting/republishing this material for advertising
or promotional purposes, creating new collective works, for resale or
redistribution to servers or lists, or reuse of any copyrighted component
of this work in other works.
\end{minipage}%
\vspace{0.5\baselineskip}}

\maketitle
\pagestyle{plain}
\thispagestyle{plain}

\begin{abstract}
\textit{Objective:} To develop and validate \algname, an interactive framework for natural language–conditioned generation of synthetic tabular data and longitudinal EHR trajectories for critically ill patients requiring mechanical ventilation and venovenous extracorporeal membrane oxygenation (VV-ECMO), enabling scenario-based evaluation of machine learning models in high-risk cohorts.
{\textit{Methods:}} \algname{} uses an optional large language model (LLM) parser for text-driven input. For single-time-point generation, we integrate variational autoencoders (VAEs) and diffusion models conditioned on observed variables. Multiple diffusion architectures support conditional longitudinal generation. The framework is evaluated on a multi-institutional VV-ECMO cohort of 874 patients using distributional fidelity metrics, imputation accuracy, adherence to clinical constraints, downstream mortality prediction tasks, and clinician evaluation.
{\textit{Results:}} Gemma 3 4B with few-shot prompting achieves the best performance in parsing clinician prompts while maintaining a low spurious slot rate. The experimental results demonstrate strengths across IGE variants: IGE-TabDDPM and IGE-PMVAE show strong performance in single-time-point generation, whereas IGE-CSDI, IGE-FGTI, and IGE-MTSCI achieve the best performance on different longitudinal fidelity metrics. In downstream evaluations across five random seeds and three classifiers, models trained on synthetic cohorts retain predictive utility for mortality prediction on real test data. Clinician ratings further indicate that the generated trajectories are clinically plausible and consistent with the requested scenarios.
{\textit{Conclusion:}} \algname{} enables clinician-specified conditional generation of clinically coherent EHR trajectories with strong statistical fidelity and downstream predictive utility in the evaluated VV-ECMO cohort and could be extended to other clinical settings.
{\textit{Significance:}} This framework supports the development and evaluation of machine learning systems in data-sparse critical care settings and advances clinically grounded synthetic data generation for biomedical research. 
\footnote{Project page: \url{https://panxulab.github.io/InterGenEHR/}}

\end{abstract}


\section{Introduction}
In critical care settings, machine learning (ML) models have demonstrated the potential to support patient management by enabling real-time risk prediction and clinical decision support \citep{wu2023value,lee2023development}.
Electronic health records (EHRs) have become a primary data source for training and evaluating ML models in critical care.
ML models trained on EHR data have shown promise in predicting patient outcomes, detecting early warning signals, and supporting treatment recommendations \citep{edelson2024early,logan2025early,song2026recmomender}.
Despite these advances, the evaluation of specific patient scenarios in intensive care unit (ICU) settings remains challenging due to limited data availability for rare and high-risk patient cohorts, as well as the heterogeneity and complexity of patient states.
In real-world applications, clinicians often need to evaluate model performance in specific patient scenarios that are not well represented in training data.
To address data scarcity, generative models have been proposed to synthesize realistic EHR data \citep{choi2017generating, li2023generating,tian2023reliable}.
However, these approaches do not fully satisfy clinicians' needs to design customized patient scenarios based on user-specified clinical features.
Therefore, interactive frameworks are needed to let clinicians explicitly define patient characteristics and generate scenario-consistent synthetic EHR data.

InterGenEHR \citep{yang2025interactive} was previously proposed as an interactive framework for generating synthetic EHR data conditioned on user-specified clinical features. 
Using VAE-based models with arbitrary conditioning \citep{ivanov2018variational,strauss2022posterior}, clinicians can define constraints on variables such as the PaO\textsubscript{2}/FiO\textsubscript{2} ratio and ventilator settings through a web interface and iteratively refine outputs via human feedback. 
This framework enables clinicians to construct patient scenarios for evaluating ML models trained on the same dataset for tasks such as mortality prediction, diagnosis prediction, and treatment decision support.
While effective, the prior approach requires manual variable selection and precise numerical specification.
The high-dimensional nature of longitudinal EHR data makes manual constraint specification difficult for clinicians, creating a barrier to targeted scenario exploration.
LLMs can reduce manual feature specification by translating clinical language into structured inputs.
Existing studies have used LLMs to extract structured variables or labels from clinical text \citep{ntinopoulos2025large,li2023two}, directly generate synthetic clinical text or tabular records \citep{li2023two,barr2025large}, and support patient-level EHR understanding and querying \citep{wu2024instruction,shi2024ehragent,ryu2024ehr}.
In \algname{}, the LLM is used only as an optional parser.
It converts a clinician's request into a JSON conditioning specification, while a separate VAE- or diffusion-based backbone generates the numerical EHR values and trajectories.
InterGenEHR uses VAE-based backbones and supports arbitrary conditioning only for records at a single time point.
It therefore does not model relationships between time points in a patient trajectory. Diffusion models offer another approach to EHR generation and can be used for either single-time-point records or longitudinal trajectories.
EHRDiff generates unconditional static binary patient records \citep{yuan2023ehrdiff}, while MedDiff supports class-conditional generation of static EHR records \citep{he2023meddiff}.
TabDDPM has also been evaluated for mixed-type static tabular EHR synthesis \citep{ceritli2023synthesizing}.
These methods still generate records at a single time point and do not model relationships between time points.
Other diffusion work generates complete longitudinal EHR trajectories with mixed-type variables \citep{kuo2023synthetic}, but it does not support arbitrary clinician-specified conditioning over features and time.

We propose \algname{} to address these limitations by connecting an LLM parsing interface with established VAE- and diffusion-based generative backbones. First, the LLM interface translates a clinician's natural-language description into a structured specification of supported observed features and values.
The framework checks the JSON format and value ranges and requires clinician confirmation before generation.
Clinicians may also bypass the LLM and enter the same constraints through the structured interface.
Second, we include diffusion-based methods in addition to the VAE-based backbones used in InterGenEHR.
These conditional backbones are adopted from prior work and connected to the same arbitrary-conditioning interface.
Finally, we extend arbitrary conditioning from single-time-point generation to longitudinal generation using a feature-by-time mask.
This extension allows the longitudinal backbones to model relationships between time points rather than generating each time point independently. 
We evaluate the LLM parsing interface and the generative framework using a pooled cohort of 874 patients receiving venovenous extracorporeal membrane oxygenation (VV-ECMO) at Duke and Emory.
The experiments on pooled patient data show that the framework can handle data combined from multiple institutions.
The contributions of this work are as follows:
\begin{itemize}[nosep,leftmargin=*]
    \item We propose \algname{}, an interactive framework that accepts either a natural-language request through an optional LLM parser or directly entered constraints. After clinician confirmation of the observed features, established VAE- and diffusion-based backbones generate the remaining EHR values for clinician-specified scenarios at a single time point or across a longitudinal trajectory.

    \item We develop an optional LLM-based semantic parsing module that translates natural-language clinical descriptions into structured conditioning specifications. The module extracts requested variables, values or ranges, and temporal information. The framework validates the JSON format, flags values outside the feature dictionary ranges, and requires clinician correction before generation.

    \item We integrate multiple conditional generative models tailored for tabular and time-series EHR data, enabling synthetic patient trajectory generation conditioned on arbitrary clinician-specified features.
    Compared with prior VAE-based approaches, the diffusion backbones are designed to capture complex joint distributions, temporal dynamics, and physiological correlations.

    \item We evaluate \algname{} on a multi-institutional dataset of VV-ECMO patients, assessing LLM parsing accuracy and generative performance (e.g., data fidelity, physiological plausibility, and clinical utility). We also conduct a clinician evaluation of generated trajectories, showing that the framework can generate clinically plausible trajectories aligned with clinician-specified scenarios. \revisiontwo{Our evaluation focuses on VV-ECMO patients, while the framework could be extended to other clinical settings. Further adaptation and validation remain future work.}
\end{itemize}

\section{Methods}

\begin{figure*}[ht]
    \centering
    \includegraphics[width=0.9\linewidth]{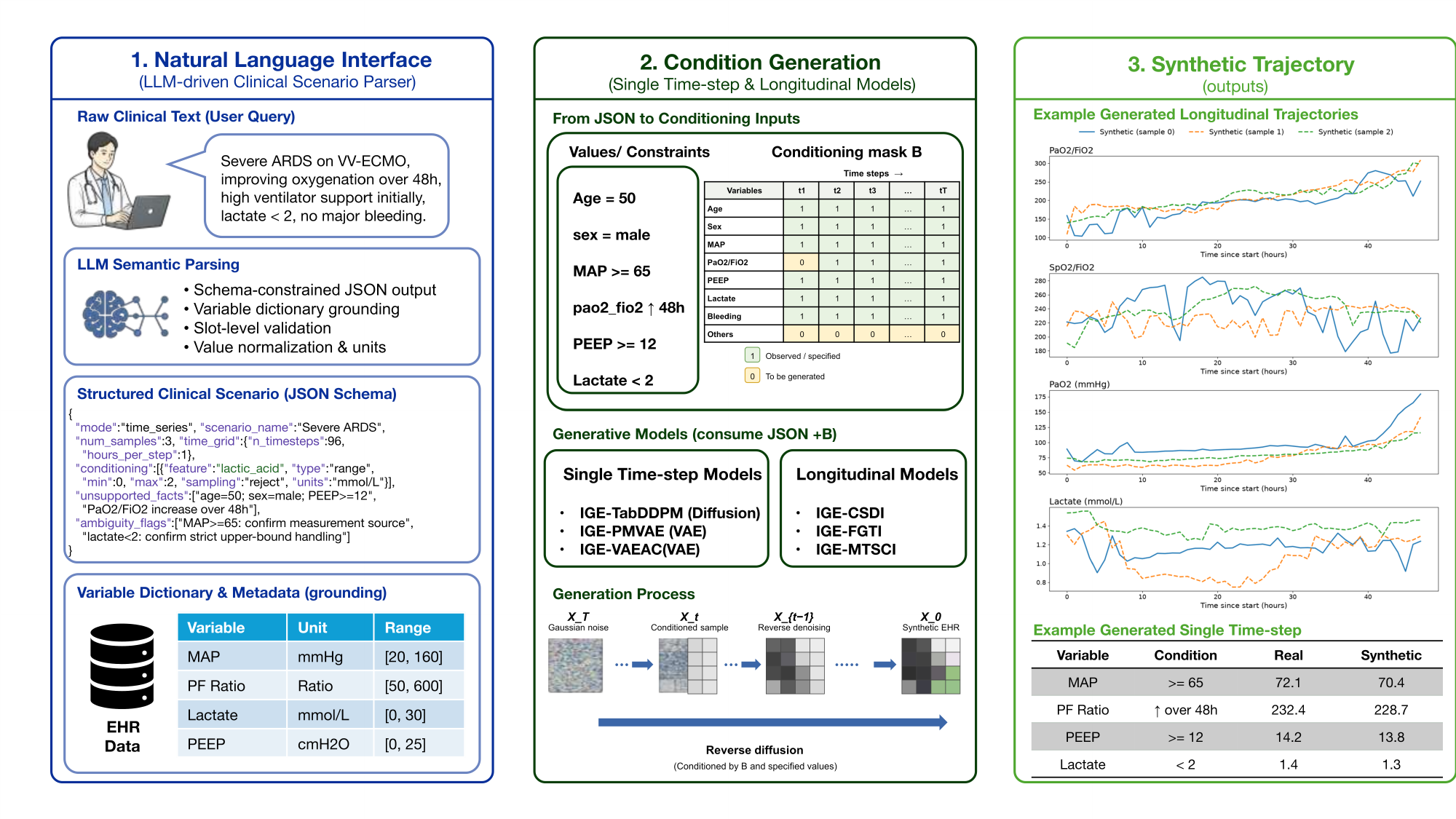}
    \caption{Overview of \algname{}. The LLM parser converts a clinician's natural-language request into structured JSON; a clinician may instead enter constraints directly. After validation and clinician confirmation, an adapter converts the constraints into observed values and observation masks $b$ or $B$. The selected conditional backbone keeps the observed entries fixed and generates a patient record at a single time point or a longitudinal trajectory.}
    \label{fig:overview}
\end{figure*}

\algname{} accepts either constraints entered through the structured interface or a natural-language request. For natural-language input, the optional LLM parser converts the request into JSON; direct input skips this step. In both cases, the constraints undergo schema and range validation, followed by clinician confirmation. A deterministic adapter converts the confirmed constraints into observed values, which the selected conditional backbone uses to generate unspecified EHR values. \Cref{fig:overview} summarizes this workflow. The following sections describe LLM parsing, its evaluation, and arbitrary conditioning at a single time point and across longitudinal trajectories. Implementation details and standard backbone formulations are provided in Appendix I and Appendix III (see Supplementary Material).

\subsection{LLM Prompt Engineering and Evaluation Framework}
\label{sec:LLM_module}
To reduce the burden of specifying conditioning variables and numeric constraints, we implement an optional LLM parser that maps clinicians' natural language into a JSON conditioning specification.
We develop a prompt engineering framework to (i) enforce structured outputs, (ii) compare alternative prompt designs, and (iii) quantify performance across models and decoding settings.

\subsubsection{Target conditioning schema and validation}
We define fixed, machine-verifiable JSON schemas for requests at a single time point and longitudinal requests. Both schemas represent a scenario name, the number of requested samples, and a list of conditioning items. Each conditioning item identifies a supported feature and a constraint type, including a value, range, or trajectory. Depending on the constraint type, the item records fields such as a value, minimum and maximum bounds, units, a trajectory pattern, endpoint values, and a time window. Model outputs are parsed and validated against the applicable schema using JSON Schema validation. Invalid JSON or output that does not follow the schema is rejected. The feature dictionary maps clinical terms to supported features and records their units and value ranges. Values outside the defined ranges are flagged for clinician review, and the clinician must confirm or modify the parsed constraints before generation. Constraints entered directly through the structured interface follow the same validation and review steps.
To identify a simple prompt design for reliable clinical parsing, we evaluate four configurations: (1) Direct Zero-Shot, (2) Direct Few-Shot, (3) Stepwise Few-Shot, and (4) Role-Anchored Few-Shot. All configurations include the target JSON schema and the feature dictionary, which enumerates supported EHR variables and their clinical semantics.

Direct Zero-Shot instructs the model to convert the clinician’s request into JSON using the schema and feature dictionary. Direct Few-Shot adds labeled examples that map clinical language to structured fields. Role-Anchored Few-Shot is identical to Direct Few-Shot but adds an expert role instruction to the system prompt to test whether specifying an expert role improves structured output reliability. Stepwise Few-Shot further adds procedural instructions that decompose the task into extracting temporal context, mapping variables, and determining constraint types before JSON construction. This design isolates the effects of demonstrations, task decomposition, and role conditioning while holding the schema constant.

\subsubsection{Benchmark set and experimental setup}
We curate a benchmark of clinician utterances paired with gold references that instantiate the target schema. Experiments are conducted across the evaluated LLM backends at decoding temperatures of 0.0, 0.2, and 0.5. These temperatures cover deterministic to moderate sampling because the task requires stable JSON rather than diverse text. For reproducibility, each run uses fixed random seeds and writes a complete execution trace, including prompts, raw model outputs, parsed JSON, and validation outcomes.

\subsubsection{Automatic evaluation metrics}
We compute both syntactic and semantic metrics. Syntactic metrics include the parse success rate and schema compliance rate. Semantic evaluation is performed at the slot level by flattening nested JSON fields into key--value ``slots'' and computing agreement with the gold reference (true positives, false positives, false negatives), from which we derive slot-level Precision, Recall, and F1 Score. We additionally report the spurious slot rate, defined as the fraction of predicted slots absent from the gold reference. This metric does not measure factual errors in free-form text. We also report an ambiguity-handling indicator based on whether the output explicitly flags uncertainty in its free-text notes.

\subsection{Single-Time-Point Arbitrary Conditioning}

We generate a patient record at a single clinical time point conditioned on observed features. Let $x\in\mathbb{R}^{d}$ denote a patient record, where $d$ is the number of clinical features. The observed-value vector is $x_o=b\odot x$, where $b\in\{0,1\}^{d}$ is a binary mask and $b_i=1$ indicates that feature $i$ is specified by the clinician. The unobserved-value vector is $x_u=(1-b)\odot x$. The goal is to estimate $p(x_u\mid x_o,b)$ and generate $\hat{x}_u$ while keeping $x_o$ fixed. The complete generated record is
$\hat{x}=x_o+(1-b)\odot\hat{x}_u
$.

We use IGE-VAEAC, IGE-PMVAE, and IGE-TabDDPM for single-time-point generation. IGE-VAEAC uses the conditional prior of VAEAC \citep{ivanov2018variational} to generate values under different observation masks. IGE-PMVAE uses the posterior-matching objective of PMVAE \citep{strauss2022posterior} to align the posteriors inferred from partial and complete observations. IGE-TabDDPM uses TabDDPM \citep{kotelnikov2023tabddpm} with replacement-based conditional sampling \citep{lugmayr2022repaint}, which keeps the observed entries fixed while generating the remaining entries. Their detailed objectives and sampling equations are provided in Appendix III.

\subsection{Longitudinal Arbitrary Conditioning}
\label{sec:time_series_generation}

We extend arbitrary conditioning from a single clinical time point to a longitudinal trajectory. Let $X\in\mathbb{R}^{d\times L}$ denote a patient trajectory, where $d$ is the number of clinical features and $L$ is the number of clinical time points. \revisiontwo{We use $t$ for clinical time and $\tau$ for diffusion steps.} The mask $B\in\{0,1\}^{d\times L}$ satisfies $B_{i,t}=1$ when feature $i$ at clinical time $t$ is specified by the clinician. The observed matrix is $X_o = B \odot X$, and the remaining values are denoted by $X_u$. The goal is to estimate $p(X_u\mid X_o,B)$ and generate the unobserved values while keeping $X_o$ fixed. The complete generated trajectory is
$\hat{X}=X_o+(1-B)\odot\hat{X}_u.$

We use IGE-CSDI, IGE-FGTI, and IGE-MTSCI for longitudinal generation. IGE-CSDI uses CSDI \citep{tashiro2021csdi} to model temporal and cross-feature relationships. IGE-FGTI uses FGTI \citep{yang2024frequencyaware} to incorporate dominant-frequency information. IGE-MTSCI uses MTSCI \citep{zhou2024mtsci} to improve consistency across adjacent trajectory segments. All three methods receive the same $X_o$ and $B$, keep the observed features fixed, and generate the remaining trajectory. Their detailed formulations are provided in Appendix III (see Supplementary Material).

\section{Results}
\label{sec:results}
We present experimental results for each component of the framework. We evaluate (1) LLM-based parsing of clinical requests, (2) single-time-point generation with VAE- and diffusion-based methods, (3) longitudinal generation with diffusion-based methods, and (4) clinical plausibility and request adherence through a structured clinician evaluation. The LLM parsing results are in Appendix V-A, and results on clinically salient features are in Appendix V-B (see Supplementary Material). \revisiontwo{We estimate 95\% confidence intervals using 1000 bootstrap samples for fidelity, data perturbation, and imputation experiments.}

\subsection{Dataset}
We collected data from adult patients who underwent VV-ECMO at two tertiary academic centers. The Emory University Hospital cohort included 183 patients treated between 2015 and 2021, while the Duke University cohort included 691 patients treated between 2016 and 2024. The dataset includes patients with documented VV-ECMO cannulation, including recorded initiation and removal dates, as well as hospital outcomes (in-hospital mortality or discharge status). 

In collaboration with clinical experts, we formatted the patient data into clinical trajectories from VV-ECMO cannulation to decannulation. We selected patients with sufficient physiological and ventilatory data during VV-ECMO support. Each trajectory was segmented with a 4-hour sliding window and a 1-hour step size. Following \citep{peine2021development}, measurements within each 4-hour window were aggregated using the median value, reducing transient fluctuations and measurement noise while preserving fine-grained temporal resolution through the 1-hour stride. Preprocessing yielded 93,149 timestamps for the Emory cohort and 295,540 timestamps for the Duke cohort. The dataset includes 35 features representing the patient's physiological state across four domains: vital signs, ventilation settings and respiratory measures, laboratory tests, and sedation scores.

To address missing values, we used forward-filling to propagate the most recent valid measurement for each patient. Residual missing values were then imputed using the population median for each variable, yielding a complete dataset for analysis.

\begin{table}[!t]
\centering
\caption{Study Cohort Demographics}
\label{tab:demographics}
\setlength{\tabcolsep}{4pt}
\resizebox{\linewidth}{!}{%
\begin{tabular}{lccc}
\toprule
 & \textbf{Overall} & \textbf{Duke} & \textbf{Emory} \\
\midrule
n & 874 & 691 & 183 \\
Age, mean (SD) & 47.8 (14.6) & 48.4 (14.8) & 45.4 (13.7) \\
\midrule
\multicolumn{4}{l}{\textit{Gender, n (\%)}} \\
\quad Female, n (\%) & 381 (43.6) & 304 (44.0) & 77 (42.1) \\
\quad Male, n (\%) & 493 (56.4) & 387 (56.0) & 106 (57.9) \\
\midrule
\multicolumn{4}{l}{\textit{Race, n (\%)}} \\
\quad African American or Black & 240 (27.5) & 168 (24.3) & 72 (39.3) \\
\quad Asian & 24 (2.7) & 19 (2.7) & 5 (2.7) \\
\quad Caucasian or White & 495 (56.6) & 429 (62.1) & 66 (36.1) \\
\quad Other/Unavailable & 115 (13.2) & 75 (10.9) & 40 (21.9) \\
\midrule
\multicolumn{4}{l}{\textit{Sequential Organ Failure Assessment scores, mean (SD)}} \\

\quad Renal, mean (SD) & 0.8 (1.0) & 0.8 (1.0) & 0.9 (1.1) \\
\quad Cardiovascular, mean (SD) & 1.9 (1.9) & 1.9 (1.9) & 2.2 (1.8) \\
\quad Respiratory, mean (SD) & 1.4 (1.4) & 1.2 (1.3) & 2.4 (1.2) \\
\midrule
Daily weight (kg), mean (SD) & 97.3 (30.5) & 93.4 (27.4) & 107.8 (35.6) \\
Mortality in hospital, n (\%) & 315 (36.0) & 255 (36.9) & 60 (32.8) \\
ECMO duration (days), mean (SD) & 16.3 (23.0) & 12.6 (19.4) & 30.3 (29.3) \\
\bottomrule
\end{tabular}}
\end{table}

\subsection{Generative Backbones and Baselines}

We summarize the generative backbones integrated into the framework and the baseline models used for comparison. Because the framework must generate unobserved features conditioned on arbitrary clinician-provided observations, we select backbones designed for conditional imputation.

\paragraph{Single-Time-Point Generation}
For single-time-point generation, we incorporate VAEAC \citep{ivanov2018variational}, PMVAE \citep{strauss2022posterior}, and TabDDPM \citep{kotelnikov2023tabddpm} as generative backbones. We denote the corresponding framework variants as \algnameAbbrv-VAEAC, \algnameAbbrv-PMVAE, and \algnameAbbrv-TabDDPM, respectively. 
We compare them with standard unconditional generative baselines, including a VAE \citep{kingma2013auto} and CTGAN \citep{xu2019modeling}. 

We also include EHRDiff \citep{yuan2023ehrdiff} and TabDiff \citep{shi2025tabdiff} as recent diffusion baselines. We adapt both methods to the continuous clinical data and single-time-point conditioning setting used in our experiments. We keep their training objectives unchanged and extend their sampling interfaces to preserve the observed values while generating the remaining features. We evaluate all methods using the same data split, conditioning masks, evaluation metrics, and preprocessing fitted on the training set. Because neither model generates longitudinal trajectories or accepts feature-by-time masks, we include them only in the single-time-point evaluations.

\paragraph{Longitudinal Generation}
For longitudinal generation, we integrate CSDI \citep{tashiro2021csdi}, FGTI \citep{yang2024frequencyaware}, and MTSCI \citep{zhou2024mtsci}. As discussed in \Cref{sec:time_series_generation}, these architectures support arbitrary conditioning and are therefore compatible with LLM-parsed constraints. We denote the corresponding variants as \algnameAbbrv-CSDI, \algnameAbbrv-FGTI, and \algnameAbbrv-MTSCI.

\subsection{Evaluation of Single-Time-Point Generation}
\label{sec:synthetic_data}
We first evaluate the fidelity of single-time-point synthetic data generated by the framework and compare it with data generated by baseline generative models. We use Maximum Absolute Mean Difference (MAMD), Root Mean Square Percentage Error (RMSPE), and Mean Absolute Percentage Error (MAPE), computed on unnormalized data. \revisiontwo{Detailed metric definitions are provided in Appendix VII (see Supplementary Material).}

These metrics quantify deviations between synthetic and real data. The results are summarized in \Cref{tab:synthetic_quality}. \algnameAbbrv-TabDDPM achieves the lowest RMSPE and MAPE. This indicates that the diffusion-based architecture generates feature values that closely align with the real data.
TabDiff achieves the lowest MAMD of $22.5158$, followed by IGE-PMVAE at $23.2106$. \revisiontwo{The high MAMD for IGE-VAEAC occurs in unconditional generation, where no observed features are provided. This setting differs from the main use of VAEAC, which is designed to generate missing values using observed features.}
Overall, these results show that diffusion and posterior-matching backbones can generate synthetic data similar to the original clinical records. EHRDiff and TabDiff are competitive in unconditional generation, but no method achieves the lowest value for every metric.

\begin{table}[!t]
\centering
\caption{\revisiontwo{Fidelity of synthetic data for single-time-point generation.}}
\label{tab:synthetic_quality}
\small
\setlength{\tabcolsep}{3pt}
\renewcommand{\arraystretch}{1}
\revisiontwo{%
\resizebox{\columnwidth}{!}{%
\begin{tabular}{@{}lccc@{}}
\toprule
 Method       & MAMD $\downarrow$     & RMSPE  $\downarrow$  & MAPE $\downarrow$    \\
\midrule
VAE     & $34.6941 \pm 7.8506$ & $10.9585 \pm 1.3814$ & $5.4688 \pm 0.4161$ \\
CTGAN   & $26.2801 \pm 8.5311$ & $15.6072 \pm 1.6838$ & $6.3271 \pm 0.4877$ \\
EHRDiff & $33.5905 \pm 14.0951$ & $11.2569 \pm 1.6008$ & $5.3466 \pm 0.6437$ \\
TabDiff & $\mathbf{22.5158} \pm 11.2001$ & $11.9823 \pm 1.9523$ & $5.4722 \pm 0.8013$ \\
IGE-PMVAE   & $23.2106 \pm 4.1469$ & $13.6258 \pm 1.7125$ & $5.9493 \pm 0.5539$ \\
IGE-VAEAC   & $136.2840 \pm 1.1137$ & $42.6889 \pm 0.2145$ & $26.9601 \pm 0.0763$ \\
IGE-TabDDPM & $33.0056 \pm 6.6981$ & $\mathbf{8.4410} \pm 1.3015$ & $\mathbf{4.8103} \pm 0.5815$ \\
\bottomrule
\end{tabular}%
}
}
\end{table}

\subsection{Evaluation of Longitudinal Generation}
We next evaluate the fidelity of generated longitudinal trajectories against real patient records. We use metrics that cover marginal distributions, discriminative quality, and dependency structures. To measure marginal distribution alignment, we use the Kolmogorov-Smirnov (KS) statistic and Total Variation (TV) distance. The KS statistic measures the distance between empirical cumulative distribution functions, while TV distance quantifies divergence between empirical feature distributions.
To evaluate overall realism, we train a two-layer long short-term memory (LSTM) classifier to distinguish real from synthetic sequences. We report the \textit{Detection Score} and \textit{Detection Accuracy}; lower detection accuracy indicates that synthetic data are harder to distinguish from real data.

To assess dependency preservation, we calculate the Frobenius norm of the difference in correlation matrices. 
Specifically, we use the Difference in Pairwise Correlation Matrices (DPCM) to measure inter-feature correlation preservation and the Difference in Cross-Sequence Matrices (DCSM) to capture temporal correlations across time steps.
The results are summarized in \Cref{tab:synthetic_quality_time_series}. 
\algnameAbbrv-CSDI achieves the lowest KS Statistic of $0.2374$ and Detection Score of $0.1338$. 
This suggests that \algnameAbbrv-CSDI effectively models the overall data distribution. Additionally, \algnameAbbrv-FGTI achieves the best performance in preserving inter-feature dependencies, with the lowest DPCM score of $3.9664$. This improvement is consistent with its frequency-aware architecture, which models feature coupling in the spectral domain. \algnameAbbrv-MTSCI preserves temporal dynamics well, achieving the lowest DCSM score.

\begin{table}[!t]
\centering
\caption{\revisiontwo{Fidelity of synthetic data for longitudinal generation.}}
\label{tab:synthetic_quality_time_series}
\small
\setlength{\tabcolsep}{3pt}
\renewcommand{\arraystretch}{1}
\revisiontwo{%
\resizebox{\columnwidth}{!}{%
\begin{tabular}{lccc}
\toprule
Metric & IGE-CSDI & IGE-FGTI & IGE-MTSCI \\
\midrule
KS Statistic $\downarrow$
& $\mathbf{0.2374} \pm 0.0075$ & $0.2393 \pm 0.0139$ & $0.3430 \pm 0.0069$ \\
TV Distance $\downarrow$
& $\mathbf{0.3108} \pm 0.0233$ & $0.3157 \pm 0.0171$ & $0.4097 \pm 0.0323$ \\
Detection Score $\downarrow$
& $\mathbf{0.1338} \pm 0.1180$ & $0.1901 \pm 0.1862$ & $0.3732 \pm 0.0807$ \\
Detection Accuracy $\downarrow$
& $\mathbf{0.6338} \pm 0.1180$ & $0.6901 \pm 0.1862$ & $0.8732 \pm 0.0807$ \\
DPCM $\downarrow$
& $5.8983 \pm 0.1464$ & $\mathbf{3.9664} \pm 0.1690$ & $6.3826 \pm 0.1130$ \\
DCSM $\downarrow$
& $7.8984 \pm 0.0842$ & $7.8830 \pm 0.0284$ & $\mathbf{7.4248} \pm 0.1881$ \\
\bottomrule
\end{tabular}%
}
}
\end{table}

\subsection{Evaluation of Single-Time-Point Perturbation}
\label{sec:perturbation_eval}

We assess conditional generation under perturbed observed features. Unlike standard generation tasks, this evaluation focuses on scenarios in which features are manually perturbed to simulate diverse patient states. We simulate clinician feedback by explicitly modifying a subset of observed features $x_o$.
The results are summarized in \Cref{tab:perturbed_results}. IGE-PMVAE achieves the lowest RMSPE of $2.3474$ and MAPE of $1.5500$, while EHRDiff achieves the lowest MAMD of $5.0932$. No method has the lowest value for every metric. This indicates that \algnameAbbrv-PMVAE captures conditional dependencies needed to generate realistic responses to specific clinical perturbations. Although VAE has a lower MAMD than IGE-PMVAE, its RMSPE and MAPE are higher than those of IGE-PMVAE.

\begin{table}[!t]
\centering
\caption{\revisiontwo{Performance comparison under single-time-point perturbation.}}
\label{tab:perturbed_results}
\small
\setlength{\tabcolsep}{3pt}
\renewcommand{\arraystretch}{1}
\revisiontwo{%
\begin{tabular}{@{}lccc@{}}
\toprule
Method & MAMD $\downarrow$ & RMSPE  $\downarrow$ & MAPE  $\downarrow$ \\
\midrule
VAE          & $6.2785 \pm 1.7785$ & $3.7036 \pm 0.2654$ & $1.8300 \pm 0.1245$ \\
EHRDiff & $\mathbf{5.0932} \pm 1.1774$ & $3.5755 \pm 0.1821$ & $1.7857 \pm 0.0776$ \\
TabDiff & $25.1226 \pm 1.2632$ & $5.4982 \pm 0.1949$ & $3.0258 \pm 0.0700$ \\
IGE-PMVAE   & $21.9434 \pm 0.9525$ & $\mathbf{2.3474} \pm 0.1822$ & $\mathbf{1.5500} \pm 0.1215$ \\
IGE-VAEAC   & $50.7934 \pm 1.2577$ & $10.2023 \pm 0.2256$ & $4.6343 \pm 0.0734$ \\
IGE-TabDDPM & $32.1774 \pm 2.6698$ & $8.5927 \pm 0.2980$ & $4.4500 \pm 0.1011$ \\
\bottomrule
\end{tabular}%
}
\end{table}

\subsection{Evaluation of Single-Time-Point Data Imputation}
\label{sec:data_imputation}

We further evaluate \algname{} on clinical data imputation. Clinicians may specify a subset of observed features $x_o$ through the LLM module based on domain knowledge or specific clinical scenarios. We formulate this task as the generation of the unobserved feature subset $x_u$, conditioned on the available observations $x_o$. Effective imputation indicates that the framework can generate plausible data from partial clinical inputs.

To simulate missing data during inference, we apply a random $50\%$ missingness mask to the clinical features in the test cohort. 
Each model reconstructs the missing features $x_u$ using only the retained observations $x_o$. Reconstruction fidelity is measured using Mean Squared Error (MSE) for pointwise accuracy and Negative Log-Likelihood (NLL) for the quality of the predictive density. The results are summarized in \Cref{tab:imputation_results}. They show the value of explicitly modeling the conditional distribution. IGE-VAEAC achieves the lowest MSE of $0.2918$, while TabDiff achieves the lowest NLL of $0.6429$. IGE-PMVAE and IGE-TabDDPM also have lower MSE values than the VAE baseline of $1.4482$. Architectures designed for imputation improve the accuracy and reliability of imputed clinical features.

\begin{table}[!t]
\centering
\caption{\revisiontwo{Performance Comparison on Single-Time-Point Imputation.}}
\label{tab:imputation_results}
\small
\setlength{\tabcolsep}{3pt}
\renewcommand{\arraystretch}{1}
\revisiontwo{%
\begin{tabular}{@{}lcc@{}}
\toprule
Method & MSE $\downarrow$ & NLL $\downarrow$ \\ 
\midrule
VAE          & 1.4482 $\pm$ 0.0072   & 4.6786 $\pm$ 0.0023  \\
EHRDiff & 1.4143 $\pm$ 0.0064 & 0.7072 $\pm$ 0.0032 \\
TabDiff & 1.2859 $\pm$ 0.0063 & \textbf{0.6429} $\pm$ 0.0032 \\
IGE-PMVAE   & 0.4227 $\pm$ 0.0054   & 0.8118 $\pm$ 0.0051  \\
IGE-VAEAC   & \textbf{0.2918} $\pm$ 0.0050   & 1.2065 $\pm$ 0.0045  \\
IGE-TabDDPM & 0.8960 $\pm$ 0.0075   & 0.9223 $\pm$ 0.0042  \\
\bottomrule
\end{tabular}%
}
\end{table}

\subsection{Evaluation of Longitudinal Data Imputation}
We extend the imputation analysis to longitudinal patient trajectories. Here, models reconstruct missing trajectory segments conditioned on observed features provided by clinicians. This task is more challenging than single-time-point imputation because models must preserve both inter-feature correlations and temporal dynamics. Following the same protocol, we apply $50\%$ missingness masks to time-series data and evaluate each model's ability to recover unobserved trajectory features. Reconstruction quality is measured using MSE for accuracy and NLL for the quality of the predictive density.

The results are summarized in \Cref{tab:data_imputation}. \algnameAbbrv-FGTI performs best on this imputation task, yielding both the lowest MSE of $0.0322$ and the lowest NLL of $-2.1719$. This precision highlights the advantage of its frequency-aware architecture. By decomposing the signal into frequency components, FGTI achieves highly accurate pointwise reconstructions. Meanwhile, \algnameAbbrv-CSDI remains competitive, indicating that its 2D attention mechanism models spatiotemporal correlations useful for imputation. Although \algnameAbbrv-MTSCI has a higher MSE of $0.0570$, it remains competitive for longitudinal imputation.
Overall, these results show that the diffusion-based variants capture temporal dependencies needed for clinical sequence imputation and can support reconstruction of missing data within patient trajectories.

\begin{table}[!t]
\centering
\caption{\revisiontwo{Performance comparison on longitudinal imputation.}}
\label{tab:data_imputation}
\small
\setlength{\tabcolsep}{3pt}
\renewcommand{\arraystretch}{1}
\revisiontwo{%
\begin{tabular}{lcc}
\toprule
Method & MSE $\downarrow$ & NLL $\downarrow$ \\
\midrule
IGE-CSDI
& $0.0478 \pm 0.0120$
& $2.6806 \pm 0.0615$ \\
IGE-FGTI
& $\mathbf{0.0322 \pm 0.0119}$
& $\mathbf{-2.1719 \pm 0.0592}$ \\
IGE-MTSCI
& $0.0570 \pm 0.0107$
& $7.2098 \pm 0.0783$ \\
\bottomrule
\end{tabular}%
}
\end{table}

\subsection{Evaluation of Clinically Constrained Single-Time-Point Generation} 

We evaluate the ability of the framework to generate clinically valid synthetic data that adhere to established clinical constraints. Traditional generative models approximate the joint distribution of the data without incorporating deterministic clinical laws, which can yield samples that violate clinical relationships.  In addition, sensor artifacts, preprocessing, and recording delays can cause even real patient records to deviate from theoretical constraints. To address this, the framework enables clinicians to specify critical variables through the LLM interface. 
By treating these variables as observed features $x_o$, the generative models condition on them and preserve the supplied constraints. We compare the framework with traditional baselines using three well-established clinical equations from \cite{yang2025interactive}. 
We selected these equations because they describe explicit relationships among variables related to oxygenation, ventilation, and acid-base balance during VV-ECMO support. All variables in these equations are available in both institutional datasets. Other clinically relevant relationships among the available variables are less deterministic and are therefore not included as fixed equation-based checks.
First, the pH is governed by the Henderson-Hasselbalch equation \citep{kitching2002acid}: 
\begin{align}
    \text{pH} = 6.1 + \log_{10}\bigg( \frac{\text{HCO}_3^-}{0.03 \times \text{PaCO}_2} \bigg).
\end{align}
Then, the relationship between base excess and pH is defined as \citep{juern2012base}:
\begin{align}
    \text{Base Excess} = 0.9287 \times \text{HCO}_3^- + 13.77 \times \text{pH} - 124.58.
\end{align}
Finally, the oxygen-hemoglobin dissociation curve is modeled by the Severinghaus equation \citep{severinghaus1979simple}: 
\begin{align}
    \text{SpO}_2 = \bigg( \frac{23400}{\text{PaO}_2^3 + 150 \times \text{PaO}_2} + 1 \bigg)^{-1}.
\end{align}

We calculate the Mean Absolute Error (MAE) between generated values and theoretically derived values from these equations. As shown in \Cref{tab:clinical_constraints}, standard generative models such as VAE and CTGAN can violate these constraints and produce clinically inconsistent samples. Real EHR data also exhibit non-zero MAE because of measurement and recording noise. 

\Cref{tab:clinical_constraints_pmvae} summarizes the conditional generation performance of \algname. When only a single variable such as pH is constrained, \algnameAbbrv-PMVAE and \algnameAbbrv-VAEAC generate the remaining related variables with MAE levels comparable to or better than the baselines in \Cref{tab:clinical_constraints}. This suggests that explicitly modeling the conditional distribution helps preserve clinical relationships among related variables.

\begin{table}[ht]
\centering
\caption{MAE for Clinical Constraints Across Generative Models.}
\label{tab:clinical_constraints}
\resizebox{\linewidth}{!}{
\begin{tabular}{lccc}
\toprule
Method & pH MAE $\downarrow$ & Base Excess MAE $\downarrow$ & SpO$_2$ MAE $\downarrow$ \\ 
\midrule
VAE           & 0.0934 & \textbf{1.2532} & \textbf{2.1890} \\
CTGAN        & 0.1125 & 4.0484 & 3.1240 \\
EHRDiff & 0.0949 & 1.5325 & 2.6732 \\
TabDiff & 0.0928 & 1.4892 & 2.3519 \\
\algnameAbbrv-PMVAE   & 0.1069 & 1.6659 & 2.4445 \\
\algnameAbbrv-VAEAC   & 0.1026 & 1.6991 & 3.1950 \\ 
\algnameAbbrv-TabDDPM & \textbf{0.0868} & 1.3772 & 2.2346 \\
\midrule
\algnameAbbrv-PMVAE (Observed)   & 0 & 0 & 0 \\
\algnameAbbrv-VAEAC (Observed)   & 0 & 0 & 0 \\
\algnameAbbrv-TabDDPM (Observed) & 0 & 0 & 0 \\
\midrule
Real Data               & 0.0834 & 1.6050 & 2.7247 \\
\bottomrule
\end{tabular}}
\end{table}

\begin{table}[ht]
\centering
\caption{MAE for Clinical Constraints Under Single-Variable Conditioning.}
\label{tab:clinical_constraints_pmvae}
\resizebox{\linewidth}{!}{
\begin{tabular}{lccc}
\toprule
Conditioned Variable & pH MAE $\downarrow$ & Base Excess MAE $\downarrow$ & SpO$_2$ MAE $\downarrow$ \\ 
\midrule
\algnameAbbrv-PMVAE (pH)           & 0 & 1.6423 & 2.2905 \\
\algnameAbbrv-PMVAE (Base Excess)  & 0.1002 & 0 & 2.3124 \\ 
\algnameAbbrv-PMVAE (SpO$_2$)      & 0.0946 & 1.5431 & 0 \\ 
\midrule
\algnameAbbrv-VAEAC (pH)           & 0 & 0.9386 & 3.6624 \\
\algnameAbbrv-VAEAC (Base Excess)  & 0.0989 & 0 & 3.7803 \\ 
\algnameAbbrv-VAEAC (SpO$_2$)      & 0.0967 & 0.9563 & 0 \\
\bottomrule
\end{tabular}}
\end{table}

\subsection{Clinician Evaluation}
\label{sec:clinician_evaluation}

\revisiontwo{In this section, we evaluate the generated VV-ECMO trajectories through direct assessment by four clinicians. For this experiment, we combine the Duke and Emory cohorts and train the IGE-FGTI model. The clinician evaluation uses the same 35 clinical variables as the main experiments.}

For each requested clinical scenario, clinicians review the generated trajectories and rate each trajectory using five questions. \revisiontwo{The 10 scenarios cover severe lung injury, persistent low blood oxygen levels, respiratory and metabolic acidosis, recovery and possible removal of ECMO support, suspected breakdown of red blood cells, a sudden drop in blood oxygen levels, difficulty clearing carbon dioxide, acute kidney injury, and falling platelet counts.} The questions assess clinical plausibility, internal physiological consistency, requested-scenario representation, physiological possibility, and \revisiontwo{agreement with the requested scenario. The complete questions and scenario definitions are provided in Appendix VI (see Supplementary Material)}. The trajectories are presented in randomized order without disclosing the generating model, and all ratings use a 1--5 scale. Higher scores indicate greater clinical plausibility or closer alignment with the requested scenario.

\revisiontwo{For each evaluation question and scenario, we compare all six pairs of scores from the four clinicians. We define ``Exact pairwise agreement'' as the proportion of pairs with identical scores (\Cref{tab:clinician_eval_results}). To allow for variation between raters, we also report ``Pairwise difference $\leq 1$'', the proportion of pairs whose scores differ by at most one point.
The mean scores range from 3.68 to 4.20 on the 1--5 scale, with higher mean ratings for clinical plausibility and physiological possibility. These results suggest that the generated trajectories are clinically plausible and broadly aligned with clinician-specified scenarios.}

\begin{table}[ht]
\centering
\revisiontwo{%
\caption{Clinician Evaluation Results with Four Independent Raters}
\label{tab:clinician_eval_results}
\resizebox{\linewidth}{!}{%
\begin{tabular}{@{}lccc@{}}
\toprule
Evaluation item
& \shortstack{Score}
& \shortstack{Exact \\ pairwise agreement}
& \shortstack{Pairwise \\ difference $\leq 1$} \\
\midrule
Q1 Clinical plausibility
& $4.10 \pm 0.59$ & 51.7\% & 91.7\% \\
Q2 Internal physiological consistency
& $3.90 \pm 0.63$ & 43.3\% & 86.7\% \\
Q3 Requested-scenario representation
& $3.80 \pm 0.61$ & 53.3\% & 93.3\% \\
Q4 Physiological possibility
& $4.20 \pm 0.65$ & 38.3\% & 93.3\% \\
Q5 Agreement with the requested scenario
& $3.68 \pm 0.62$ & 46.7\% & 91.7\% \\
\bottomrule
\end{tabular}%
}%
}
\end{table}

\subsection{Downstream Clinical Utility: In-Hospital Mortality Prediction}

\begin{figure}[ht]
\centering
\includegraphics[width=0.48\textwidth]{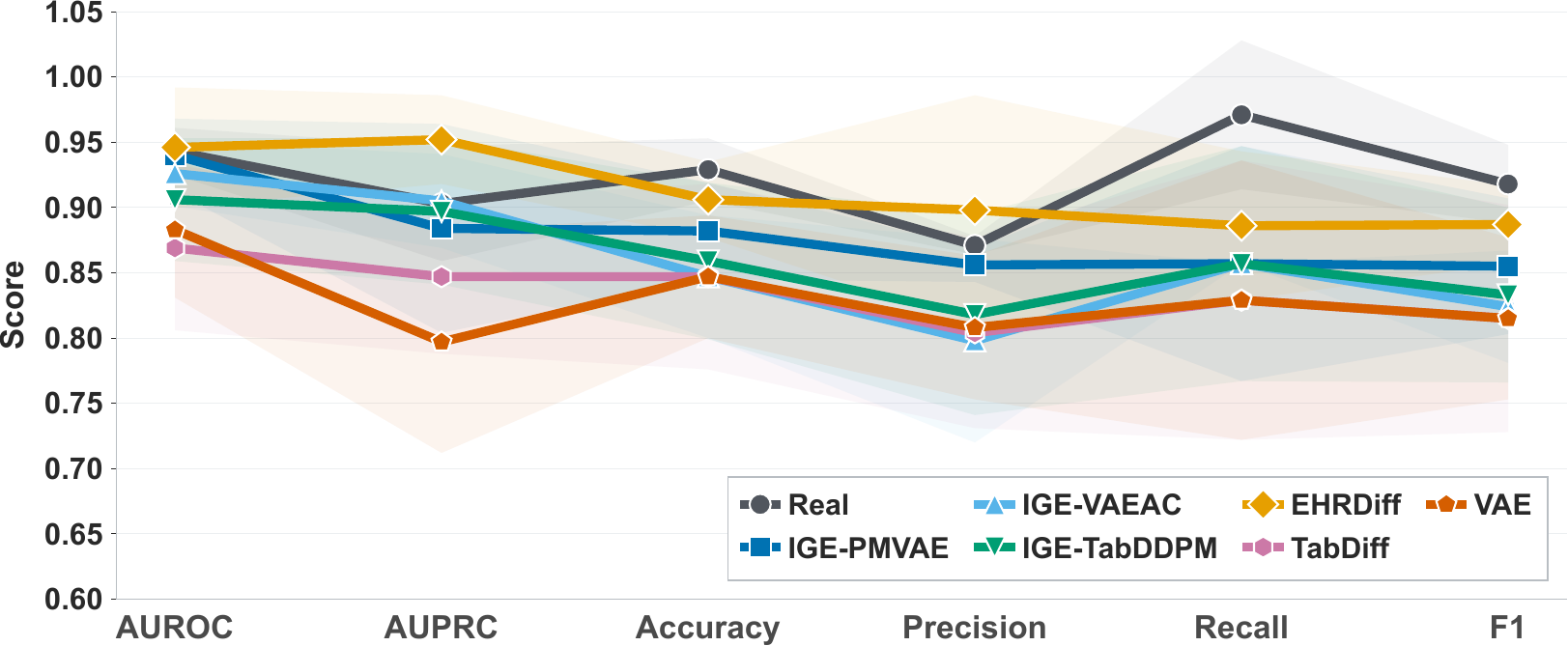}
\caption{\revisiontwo{MLP-based in-hospital mortality classification results under unconditional generation without clinician-provided observed features.}}
\label{fig:mlp_classification_results}
\end{figure}

\begin{table*}[!t]
\centering
\caption{\revisiontwo{MLP-based in-hospital mortality classification results using scenario-conditioned synthetic data.}}
\label{tab:scenario10_mlp_tstr}
\small
\setlength{\tabcolsep}{3pt}
\renewcommand{\arraystretch}{1}
\revisiontwo{%
\begin{tabular}{@{}lcccccc@{}}
\toprule
Training data & AUROC & AUPRC & Accuracy & Precision & Recall & F1 \\
\midrule
IGE-PMVAE & $0.7631 \pm 0.0521$ & $0.7778 \pm 0.0879$ & $0.7697 \pm 0.0873$ & $0.7743 \pm 0.1904$ & $0.6615 \pm 0.0877$ & $0.6985 \pm 0.0827$ \\
IGE-TabDDPM & $0.7223 \pm 0.0320$ & $0.7237 \pm 0.0413$ & $0.6667 \pm 0.0479$ & $0.5933 \pm 0.0760$ & $0.5077 \pm 0.1032$ & $0.5427 \pm 0.0748$ \\
IGE-VAEAC & $0.6123 \pm 0.1334$ & $0.6321 \pm 0.0767$ & $0.6545 \pm 0.0819$ & $0.6218 \pm 0.1682$ & $0.4308 \pm 0.0877$ & $0.4971 \pm 0.0789$ \\
IGE-CSDI & $0.7638 \pm 0.0548$ & $0.7340 \pm 0.0820$ & $0.7091 \pm 0.0507$ & $0.6830 \pm 0.1099$ & $0.5077 \pm 0.1685$ & $0.5662 \pm 0.1326$ \\
IGE-FGTI & $0.6138 \pm 0.1507$ & $0.6729 \pm 0.1256$ & $0.7455 \pm 0.0730$ & $0.9667 \pm 0.0745$ & $0.3846 \pm 0.2433$ & $0.5088 \pm 0.2109$ \\
IGE-MTSCI & $0.7523 \pm 0.0413$ & $0.7678 \pm 0.0279$ & $0.7333 \pm 0.0332$ & $0.7575 \pm 0.1135$ & $0.5385 \pm 0.1804$ & $0.6047 \pm 0.0690$ \\
\bottomrule
\end{tabular}%
}
\end{table*}

We evaluate the utility of the generated synthetic data through a critical downstream clinical task: predicting in-hospital mortality. High-fidelity synthetic data should preserve the underlying predictive relationships between clinical features and patient outcomes, thereby serving as an augmentation tool for real patient data in machine learning pipelines. 

We focus on the VV-ECMO dataset, extracting time-series trajectories and their corresponding in-hospital mortality labels. To evaluate the utility of the generated data, we employ the ``Train on Synthetic, Test on Real'' (TSTR) paradigm. We train three classifiers---logistic regression (LR), gradient-boosted decision trees (GBDT), and a multilayer perceptron (MLP)---on the synthetic data and evaluate them on real test data. We report train-on-real, test-on-real performance as the reference. Each experiment is repeated with five seeds, and we report the mean and standard deviation for AUROC, AUPRC, accuracy, precision, recall, and F1 score.
\revisiontwo{This reference evaluation uses unconditional generation without clinician-provided observed features or scenario-specific constraints.} Overall, IGE-PMVAE has the highest mean AUROC among the synthetic-data methods with LR and GBDT, while EHRDiff has the highest mean AUROC with MLP classifiers. \revisiontwo{Complete numerical results for MLP, LR, and GBDT under unconditional generation are provided in Appendix V-C (see Supplementary Material).} \revisiontwo{For illustration, the MLP results are shown in \Cref{fig:mlp_classification_results}.} Performance varies across generative methods and classifiers; synthetic data may not fully preserve all predictive information. \revisiontwo{Although EHRDiff achieves the highest mean MLP AUROC, its performance is close to the real-data reference.}

\revisiontwo{We further evaluate scenario-conditioned generation using the ten clinician-specified scenarios in Appendix VI. For each model, we generate 1,400 synthetic records. The clinical features specified in each parsed scenario are fixed to their observed encounter values, while the remaining features are generated. These features correspond to the clinician-requested conditions. We train an MLP classifier and evaluate it via TSTR. As shown in Table~\ref{tab:scenario10_mlp_tstr}, IGE-CSDI and IGE-PMVAE achieve the highest AUROC values of 0.7638 and 0.7631, respectively, indicating that the synthetic data retain predictive utility.}

\section{Discussion}
In this work, we introduce \algname, an interactive framework that integrates LLM-based semantic parsing with conditional generative modeling to enable clinician-driven synthesis of tabular and longitudinal EHR trajectories. By combining structured natural language interpretation with diffusion and posterior-matching architectures, the proposed system reduces the manual burden of feature specification while supporting conditional generation both at single time points and across longitudinal trajectories. Across multi-institutional VV-ECMO cohorts, we show that LLM prompting strategies substantially affect structured parsing reliability, and different diffusion and posterior-matching variants show strengths on different fidelity, conditioning, and imputation metrics.

The LLM evaluation shows that prompt design affects the reliability of clinical scenario parsing. Zero-shot prompting produces lower semantic accuracy and schema compliance, whereas few-shot prompting achieves the highest slot-level F1 and lowest spurious slot rates. These results suggest that providing examples is more effective than adding stepwise instructions for this structured parsing task. The feature dictionary and JSON schema further constrain the model outputs and support validation before generation.

\revisiontwo{Schema validation cannot detect all parsing errors. LLMs may add information not supported by clinicians' requests (spurious slots) or generate incorrect values. For example, a parser may give a creatinine value of $150$~mg/dL, above the $20$~mg/dL upper limit in our feature dictionary. Such errors can distort generated trajectories and mislead downstream analysis. The framework checks value ranges and feature names, flags out-of-range values for review and correction, and excludes unsupported features from conditioning. Only conditions confirmed by clinicians are used for generation.}

For single-time-point generation, diffusion and posterior-matching approaches exhibit complementary strengths. The diffusion-based method \algnameAbbrv-TabDDPM achieves the lowest relative error metrics, showing better agreement with marginal feature statistics. In contrast, \algnameAbbrv-PMVAE achieves the lowest relative error metrics under single-time-point perturbation, reflecting improved performance under conditional perturbation. By explicitly aligning latent posteriors under partial observations, \algnameAbbrv-PMVAE better preserves conditional relationships when clinicians specify arbitrary subsets of features. These findings suggest that diffusion enhances fidelity in unconditional generation, while posterior matching strengthens performance under conditioning. Both are critical for interactive scenario generation. Single-time-point conditional generation also enables practical testing of ML models at specific clinical decision points. For example, it can complete partially specified records and construct rare or borderline cases to evaluate model performance in patient states underrepresented in the available data.

In the longitudinal setting, different diffusion architectures show different strengths across evaluation metrics. \algnameAbbrv-CSDI provides strong marginal distribution alignment and low detection scores, indicating effective modeling of the overall data distribution. \algnameAbbrv-FGTI achieves the best preservation of inter-feature correlations and the best imputation performance, likely due to its frequency-aware design that captures temporal patterns at different frequencies. \algnameAbbrv-MTSCI best preserves temporal correlations across time steps. No single architecture uniformly dominates across all metrics, underscoring that model selection should depend on whether marginal distribution fidelity, inter-feature correlation, temporal correlation, or imputation accuracy is prioritized.

In clinical settings, this framework enables scenario-based evaluation of machine learning systems in high-risk cohorts such as VV-ECMO patients, where rare physiological states are often underrepresented. By allowing clinicians to specify targeted trajectories, the system supports the evaluation of predictive models, policy evaluation in reinforcement learning settings, and structured data augmentation. However, synthetic plausibility does not imply causal validity. The generated trajectories reflect learned statistical structure rather than mechanistic physiology and should therefore be interpreted as simulation tools rather than clinical decision engines.

\revisiontwo{The imbalance between Duke (691 patients) and Emory (183 patients) is a limitation of this study. The generated data may therefore reflect Duke patients more strongly and may not generalize to other institutions. We did not separately evaluate site-specific effects because the smaller Emory cohort would make these comparisons less reliable. Further evaluation with larger and more balanced multi-site cohorts is needed.} \revisiontwo{Validation on external cohorts and in other clinical settings is needed to establish broader applicability.}

\section{Conclusion}
We introduce \algname, a clinician-interactive framework that combines structured LLM-based scenario parsing with conditional diffusion and posterior-matching generative models to synthesize realistic EHR trajectories. By enabling natural language specification and arbitrary conditioning, the system improves usability, fidelity, and temporal coherence in both single-time-point and longitudinal generation. Evaluation on multi-institutional VV-ECMO cohorts demonstrates reliable scenario translation and high-quality synthetic data generation, supporting simulation-driven evaluation of machine learning models in \revisiontwo{the evaluated VV-ECMO setting}.

\section{Ethics Statement}
This study was reviewed by the Duke University Health System Institutional Review Board and determined to be exempt under IRBPro00114885. All procedures were conducted in accordance with applicable regulations and institutional guidelines.

\section*{Acknowledgment}
This work was supported in part by the National Institutes of Health under Award Numbers R01GM139967 and R21GM148931 (to R. Kamaleswaran and J. Song). H. P. Foote is supported by the Eunice Kennedy Shriver National Institute of Child Health and Human Development under Awards T32HD094671 and T32HD104576. P. Xu and Y. Yang are supported by the National Science Foundation under Grant No. IIS-2541960 (CAREER) and the Whitehead Scholars Program at the Duke University School of Medicine.

\bibliographystyle{IEEEtran}
\bibliography{reference}

\appendices

\onecolumn

\section{\revision{Implementation Details}}
\label{app:implementation_details}

\revision{The following implementation details correspond to the pooled Duke--Emory experiments with 35 clinical variables.}

\paragraph{\revision{VAE}}

\revision{The VAE is used as an unconditional reference model for generation at a single time point. It maps the 35-variable input through a two-layer MLP encoder to a 32-dimensional latent representation and reconstructs the input with a symmetric decoder. The architecture and optimization settings are summarized in Table~\ref{tab:vae_hyperparameters}.}

\begin{table}[!htbp]
\centering
\caption{\revision{Hyperparameters for VAE.}}
\label{tab:vae_hyperparameters}
\footnotesize
\begin{tabular}{@{}p{0.28\columnwidth}p{0.66\columnwidth}@{}}
\toprule
\revision{Hyperparameter} & \revision{Setting} \\
\midrule
\revision{Encoder} & \revision{MLP, $35\rightarrow256\rightarrow128$} \\
\revision{Decoder} & \revision{Symmetric MLP decoder} \\
\revision{Latent dimension} & \revision{32} \\
\revision{Optimizer} & \revision{Adam} \\
\revision{Learning rate} & \revision{$1\times10^{-4}$} \\
\revision{Weight decay} & \revision{$1\times10^{-5}$} \\
\revision{Batch size} & \revision{64} \\
\revision{Training length} & \revision{Up to 300 epochs} \\
\revision{KL weight} & \revision{$\beta=0.05$} \\
\revision{Learning-rate schedule} & \revision{Cosine, $T_{\max}=50$} \\
\bottomrule
\end{tabular}
\end{table}

\paragraph{\revision{CTGAN}}

\revision{CTGAN is used as a GAN-based reference model for unconditional tabular synthesis. The generator and discriminator each contain two hidden layers, and packing is applied to the discriminator to reduce mode collapse. Table~\ref{tab:ctgan_hyperparameters} reports the adversarial optimization and training settings.}

\begin{table}[!htbp]
\centering
\caption{\revision{Hyperparameters for CTGAN.}}
\label{tab:ctgan_hyperparameters}
\footnotesize
\begin{tabular}{@{}p{0.28\columnwidth}p{0.66\columnwidth}@{}}
\toprule
\revision{Hyperparameter} & \revision{Setting} \\
\midrule
\revision{Embedding dimension} & \revision{128} \\
\revision{Generator dimensions} & \revision{$(256,256)$} \\
\revision{Discriminator dimensions} & \revision{$(256,256)$} \\
\revision{PAC size} & \revision{10} \\
\revision{Optimizer} & \revision{Adam, $\beta_1=0.5$, $\beta_2=0.9$} \\
\revision{Learning rate} & \revision{$2\times10^{-4}$ for generator and discriminator} \\
\revision{Weight decay} & \revision{$1\times10^{-6}$ for generator and discriminator} \\
\revision{Batch size} & \revision{500} \\
\revision{Training length} & \revision{300 epochs} \\
\bottomrule
\end{tabular}
\end{table}

\paragraph{\revision{EHRDiff}}

\revision{EHRDiff is trained as an unconditional EDM model on the standardized training data. The input dimension is set from the pooled training data after preprocessing. For the conditional evaluations, we add replacement-based sampling that restores the observed entries at the corresponding noise level during each reverse-diffusion step and generates the remaining entries. This sampling adaptation accepts a separate observation mask and observed values for each record without changing the EHRDiff denoiser or training objective. Table~\ref{tab:ehrdiff_hyperparameters} summarizes the configuration.}

\begin{table}[!htbp]
\centering
\caption{\revision{Hyperparameters for EHRDiff.}}
\label{tab:ehrdiff_hyperparameters}
\footnotesize
\begin{tabular}{@{}p{0.28\columnwidth}p{0.66\columnwidth}@{}}
\toprule
\revision{Hyperparameter} & \revision{Setting} \\
\midrule
\revision{Denoiser} & \revision{MLP, four hidden layers of width 256} \\
\revision{Time embedding} & \revision{128} \\
\revision{Optimizer} & \revision{AdamW} \\
\revision{Learning rate} & \revision{$3\times10^{-4}$} \\
\revision{Weight decay} & \revision{0} \\
\revision{Batch size} & \revision{256} \\
\revision{Training length} & \revision{30,000 optimization steps} \\
\revision{Exponential moving average rate} & \revision{0.999} \\
\revision{Sampling} & \revision{32-step second-order Heun sampler; $\sigma_{\min}=0.02$, $\sigma_{\max}=80$, $\rho=7$} \\
\revision{Conditional adaptation} & \revision{Replacement-based sampling with a per-record observation mask} \\
\bottomrule
\end{tabular}
\end{table}

\paragraph{\revision{TabDiff}}

\revision{TabDiff is trained on the same pooled training data using its continuous-time diffusion objective, learnable feature-wise noise schedules, and the released UniModMLP backbone with two encoder and decoder layers. Its original imputation interface uses a designated target column. We extend the conditional sampling procedure to accept a separate observation mask and observed values for each record. The replacement-based sampler restores the observed entries and generates the remaining entries. This adaptation leaves the TabDiff training objective unchanged. Table~\ref{tab:tabdiff_hyperparameters} summarizes the configuration.}

\begin{table}[!htbp]
\centering
\caption{\revision{Hyperparameters for TabDiff.}}
\label{tab:tabdiff_hyperparameters}
\footnotesize
\begin{tabular}{@{}p{0.28\columnwidth}p{0.66\columnwidth}@{}}
\toprule
\revision{Hyperparameter} & \revision{Setting} \\
\midrule
\revision{Denoiser} & \revision{UniModMLP with two encoder and decoder layers} \\
\revision{Optimizer} & \revision{AdamW} \\
\revision{Learning rate} & \revision{$1\times10^{-3}$} \\
\revision{Weight decay} & \revision{0} \\
\revision{Batch size} & \revision{4,096} \\
\revision{Training length} & \revision{8,000 optimization steps} \\
\revision{Learning-rate schedule} & \revision{Reduce on plateau, factor 0.90, patience 50} \\
\revision{Exponential moving average rate} & \revision{0.997} \\
\revision{Diffusion steps} & \revision{50} \\
\revision{Noise schedule} & \revision{Learnable feature-wise power-mean schedule} \\
\revision{Sampling} & \revision{Stochastic sampler with second-order correction} \\
\revision{Conditional adaptation} & \revision{Replacement-based sampling with a per-record observation mask} \\
\bottomrule
\end{tabular}
\end{table}

\paragraph{\revision{IGE-PMVAE}}

\revision{IGE-PMVAE extends the VAE backbone with a partial encoder that maps observed values and their mask to the latent representation required for arbitrary conditioning. The model is trained in two stages: the VAE is fitted first, followed by the partial encoder under random masking. The complete settings are provided in Table~\ref{tab:pmvae_hyperparameters}.}

\begin{table}[!htbp]
\centering
\caption{\revision{Hyperparameters for IGE-PMVAE.}}
\label{tab:pmvae_hyperparameters}
\footnotesize
\begin{tabular}{@{}p{0.28\columnwidth}p{0.66\columnwidth}@{}}
\toprule
\revision{Hyperparameter} & \revision{Setting} \\
\midrule
\revision{VAE backbone} & \revision{VAE architecture in Table~\ref{tab:vae_hyperparameters}} \\
\revision{Partial encoder} & \revision{$70\rightarrow256\rightarrow256\rightarrow128\rightarrow128$} \\
\revision{Latent dimension} & \revision{32} \\
\revision{Optimizer} & \revision{Adam} \\
\revision{VAE learning rate} & \revision{$1\times10^{-4}$} \\
\revision{VAE weight decay} & \revision{$1\times10^{-5}$} \\
\revision{Partial-encoder learning rate} & \revision{$1\times10^{-3}$} \\
\revision{Partial-encoder weight decay} & \revision{$1\times10^{-5}$} \\
\revision{Batch size} & \revision{64} \\
\revision{Training length} & \revision{300 VAE epochs followed by up to 600 partial-encoder epochs} \\
\revision{Random masking ratio} & \revision{50\%} \\
\revision{Learning-rate schedule} & \revision{Cosine, $T_{\max}=50$} \\
\revision{Early-stopping patience} & \revision{100 epochs} \\
\bottomrule
\end{tabular}
\end{table}

\paragraph{\revision{IGE-VAEAC}}

\revision{IGE-VAEAC directly learns conditional distributions for arbitrary observed-feature subsets. During training, 50\% random masks expose the model to varying conditioning patterns, and importance-weighted samples are used for validation. Table~\ref{tab:vaeac_hyperparameters} summarizes the network and training configuration.}

\begin{table}[!htbp]
\centering
\caption{\revision{Hyperparameters for IGE-VAEAC.}}
\label{tab:vaeac_hyperparameters}
\footnotesize
\begin{tabular}{@{}p{0.28\columnwidth}p{0.66\columnwidth}@{}}
\toprule
\revision{Hyperparameter} & \revision{Setting} \\
\midrule
\revision{Network width} & \revision{256} \\
\revision{Network depth} & \revision{10} \\
\revision{Latent dimension} & \revision{64} \\
\revision{Optimizer} & \revision{Adam} \\
\revision{Learning rate} & \revision{$3\times10^{-4}$} \\
\revision{Weight decay} & \revision{0} \\
\revision{Batch size} & \revision{64} \\
\revision{Training length} & \revision{300 epochs} \\
\revision{Random masking ratio} & \revision{50\%} \\
\revision{Validation samples} & \revision{25 importance-weighted samples} \\
\bottomrule
\end{tabular}
\end{table}

\paragraph{\revision{IGE-TabDDPM}}

\revision{IGE-TabDDPM generates records at a single time point with an MLP denoiser and a discrete Gaussian diffusion process. We use a cosine noise schedule and exponential moving averaging during optimization. Detailed settings are listed in Table~\ref{tab:tabddpm_hyperparameters}.}

\begin{table}[!htbp]
\centering
\caption{\revision{Hyperparameters for IGE-TabDDPM.}}
\label{tab:tabddpm_hyperparameters}
\footnotesize
\begin{tabular}{@{}p{0.28\columnwidth}p{0.66\columnwidth}@{}}
\toprule
\revision{Hyperparameter} & \revision{Setting} \\
\midrule
\revision{Denoiser} & \revision{MLP, hidden dimensions $(1024,512\times6,128)$} \\
\revision{Time embedding} & \revision{128} \\
\revision{Optimizer} & \revision{AdamW} \\
\revision{Learning rate} & \revision{$1.3926\times10^{-3}$} \\
\revision{Weight decay} & \revision{0} \\
\revision{Batch size} & \revision{4,096} \\
\revision{Training length} & \revision{100,000 optimization steps} \\
\revision{Learning-rate schedule} & \revision{Linear decay} \\
\revision{Exponential moving average rate} & \revision{0.999} \\
\revision{Diffusion steps} & \revision{1,000} \\
\revision{Noise schedule} & \revision{Cosine} \\
\bottomrule
\end{tabular}
\end{table}

\paragraph{\revision{IGE-CSDI}}

\revision{IGE-CSDI models multivariate trajectories using residual diffusion layers with temporal and feature embeddings. It is trained on 96-step trajectories and uses conditional diffusion-based denoising to generate the unobserved entries while conditioning on feature-by-time observation masks. Table~\ref{tab:csdi_hyperparameters} reports the architecture, optimization, and diffusion settings.}

\begin{table}[!htbp]
\centering
\caption{\revision{Hyperparameters for IGE-CSDI.}}
\label{tab:csdi_hyperparameters}
\footnotesize
\begin{tabular}{@{}p{0.28\columnwidth}p{0.66\columnwidth}@{}}
\toprule
\revision{Hyperparameter} & \revision{Setting} \\
\midrule
\revision{Residual layers} & \revision{4} \\
\revision{Channels} & \revision{64} \\
\revision{Attention heads} & \revision{8} \\
\revision{Time embedding} & \revision{128} \\
\revision{Feature embedding} & \revision{16} \\
\revision{Optimizer} & \revision{Adam} \\
\revision{Learning rate} & \revision{$1\times10^{-3}$} \\
\revision{Weight decay} & \revision{$1\times10^{-6}$} \\
\revision{Batch size} & \revision{16} \\
\revision{Training length} & \revision{200 epochs} \\
\revision{Learning-rate schedule} & \revision{Step decay, $\times0.1$ at epochs 150 and 180} \\
\revision{Trajectory length} & \revision{96} \\
\revision{Diffusion steps} & \revision{50} \\
\revision{Noise schedule} & \revision{Quadratic, $\beta:10^{-4}\rightarrow0.5$} \\
\bottomrule
\end{tabular}
\end{table}

\paragraph{\revision{IGE-FGTI}}

\revision{IGE-FGTI augments the longitudinal diffusion backbone with a frequency encoder to capture dominant spectral components in physiological trajectories. The implementation uses a frequency cutoff of 0.3 and retains the 10 dominant frequencies. The remaining settings are given in Table~\ref{tab:fgti_hyperparameters}.}

\begin{table}[!htbp]
\centering
\caption{\revision{Hyperparameters for IGE-FGTI.}}
\label{tab:fgti_hyperparameters}
\footnotesize
\begin{tabular}{@{}p{0.28\columnwidth}p{0.66\columnwidth}@{}}
\toprule
\revision{Hyperparameter} & \revision{Setting} \\
\midrule
\revision{Residual layers} & \revision{4} \\
\revision{Channels} & \revision{128} \\
\revision{Frequency-encoder layers} & \revision{4} \\
\revision{Attention heads} & \revision{8} \\
\revision{Time embedding} & \revision{128} \\
\revision{Feature embedding} & \revision{16} \\
\revision{Optimizer} & \revision{Adam} \\
\revision{Learning rate} & \revision{$1\times10^{-3}$} \\
\revision{Weight decay} & \revision{$1\times10^{-6}$} \\
\revision{Batch size} & \revision{16} \\
\revision{Training length} & \revision{200 epochs} \\
\revision{Learning-rate schedule} & \revision{Step decay, $\times0.1$ at epochs 150 and 180} \\
\revision{Trajectory length} & \revision{96} \\
\revision{Frequency configuration} & \revision{Cutoff 0.3; top 10 dominant frequencies} \\
\revision{Diffusion steps} & \revision{50} \\
\revision{Noise schedule} & \revision{Quadratic, $\beta:10^{-4}\rightarrow0.2$} \\
\bottomrule
\end{tabular}
\end{table}

\paragraph{\revision{IGE-MTSCI}}

\revision{IGE-MTSCI uses a residual diffusion architecture with an additional consistency objective to improve agreement across generated trajectory segments. Noise-prediction and consistency losses are jointly optimized with separately weighted terms. Table~\ref{tab:mtsci_hyperparameters} provides the complete implementation settings.}

\begin{table}[!htbp]
\centering
\caption{\revision{Hyperparameters for IGE-MTSCI.}}
\label{tab:mtsci_hyperparameters}
\footnotesize
\begin{tabular}{@{}p{0.28\columnwidth}p{0.66\columnwidth}@{}}
\toprule
\revision{Hyperparameter} & \revision{Setting} \\
\midrule
\revision{Residual layers} & \revision{4} \\
\revision{Channels} & \revision{64} \\
\revision{Attention heads} & \revision{8} \\
\revision{Time embedding} & \revision{128} \\
\revision{Feature embedding} & \revision{16} \\
\revision{Optimizer} & \revision{Adam} \\
\revision{Learning rate} & \revision{$1\times10^{-3}$} \\
\revision{Weight decay} & \revision{$1\times10^{-6}$} \\
\revision{Batch size} & \revision{16} \\
\revision{Training length} & \revision{300 epochs} \\
\revision{Learning-rate schedule} & \revision{Step decay, $\times0.1$ at epochs 150 and 225} \\
\revision{Trajectory length} & \revision{96} \\
\revision{Loss weights} & \revision{Noise $\alpha=1$; consistency $\beta=5$} \\
\revision{Diffusion steps} & \revision{50} \\
\revision{Noise schedule} & \revision{Quadratic, $\beta:10^{-4}\rightarrow0.2$} \\
\bottomrule
\end{tabular}
\end{table}

\section{\revision{LLM Prompt Templates and JSON Schemas}}
\label{app:llm_prompts_schemas}

\revision{Listings~\ref{lst:single_step_prompt} and~\ref{lst:time_series_prompt} provide the complete few-shot prompt templates for parsing requests at a single time point and across a longitudinal trajectory. The prompt-family instructions used in the LLM evaluation are provided in the LLM Module Evaluation subsection. Each evaluation prompt combines the corresponding instruction with the applicable schema and feature catalog reported here. Both templates use the 35 clinical variables listed in Table~\ref{tab:dataset_variables}, together with their units and value ranges. Listings~\ref{lst:single_step_schema} and~\ref{lst:time_series_schema} provide the corresponding JSON Schemas. The schemas validate the output fields, supported feature names, and conditioning types. Value ranges are checked separately against the feature dictionary before clinician confirmation.}

\begin{lstlisting}[
caption={\revision{Complete Few-Shot Prompt Template for Parsing a Request at a Single Time Point}},
label={lst:single_step_prompt},
basicstyle=\color{black}\ttfamily\normalsize,
breakatwhitespace=false,
columns=fullflexible
]
# SYSTEM
You convert a clinician's natural-language request (plus any explicit feature values) into ONE JSON object that drives a SINGLE-TIME-POINT synthetic-patient generator. The JSON must validate against `single_step.schema.json`.

RULES (must follow exactly):
- Output ONLY one JSON object, no prose, no markdown fences.
- `feature` MUST be copied verbatim from the catalog below. NEVER invent or abbreviate a feature name.
- Keep numeric values in the listed `unit`, within the `plausible range`, unless the clinician explicitly asks for an extreme.
- Any quantity the clinician mentions that is NOT in the catalog (e.g. FiO2, PEEP, lung compliance, plateau pressure, urine output, ventilator/ECMO settings) goes in `unsupported_facts` as a short string -- NEVER in `conditioning`.
- Put anything you could not map confidently into `ambiguity_flags`; set `confidence` (0..1) to your overall parse confidence.
- Leave any feature you are not told about OUT of `conditioning` (the generator fills it in).

# ALLOWED FEATURES (35) -- names, units, plausible ranges
| feature (use this exact name) | unit | plausible range | type |
|---|---|---|---|
| `temperature` | degC | 30 -- 43 | continuous |
| `map_line` | mmHg | 20 -- 160 | continuous |
| `map_cuff` | mmHg | 20 -- 160 | continuous |
| `pulse` | bpm | 20 -- 220 | continuous |
| `unassisted_resp_rate` | breaths/min | 0 -- 60 | continuous |
| `end_tidal_co2` | mmHg | 0 -- 80 | continuous |
| `o2_flow_rate` | L/min | 0 -- 60 | continuous |
| `base_excess` | mmol/L | -30 -- 30 | continuous |
| `bicarb_(hco3)` | mmol/L | 5 -- 50 | continuous |
| `blood_urea_nitrogen_(bun)` | mg/dL | 1 -- 200 | continuous |
| `creatinine` | mg/dL | 0.1 -- 20 | continuous |
| `phosphorus` | mg/dL | 0.5 -- 15 | continuous |
| `hemoglobin` | g/dL | 2 -- 22 | continuous |
| `met_hgb` | %
| `platelets` | 10^3/uL | 1 -- 1000 | continuous |
| `white_blood_cell_count` | 10^3/uL | 0 -- 100 | continuous |
| `carboxy_hgb` | %
| `alanine_aminotransferase_(alt)` | U/L | 0 -- 10000 | continuous |
| `ammonia` | umol/L | 0 -- 500 | continuous |
| `aspartate_aminotransferase_(ast)` | U/L | 0 -- 10000 | continuous |
| `bilirubin_total` | mg/dL | 0 -- 50 | continuous |
| `fibrinogen` | mg/dL | 30 -- 1000 | continuous |
| `inr` | ratio | 0.5 -- 12 | continuous |
| `lactate_dehydrogenase` | U/L | 50 -- 10000 | continuous |
| `lactic_acid` | mmol/L | 0 -- 30 | continuous |
| `partial_prothrombin_time_(ptt)` | sec | 10 -- 200 | continuous |
| `lipase` | U/L | 0 -- 10000 | continuous |
| `partial_pressure_of_carbon_dioxide_(paco2)` | mmHg | 10 -- 150 | continuous |
| `ph` | pH | 6.6 -- 7.8 | continuous |
| `saturation_of_oxygen_(sao2)` | %
| `procalcitonin` | ng/mL | 0 -- 500 | continuous |
| `gcs_total_score` | score | 3 -- 15 | ordinal |
| `spo2` | %
| `partial_pressure_of_oxygen_(pao2)` | mmHg | 20 -- 700 | continuous |
| `rass_score` | score | -5 -- 4 | ordinal |

# OUTPUT FORMAT
{
  "mode": "single_step",
  "scenario_name": "<short label>",
  "csn_or_template": null,
  "num_samples": <int >= 1>,
  "seed": <int or null>,
  "conditioning": [
    {"feature": "<name>", "type": "value", "value": <number>, "units": "<unit>"},
    {"feature": "<name>", "type": "range", "min": <number>, "max": <number>, "sampling": "uniform|midpoint|reject", "units": "<unit>"}
  ],
  "unsupported_facts": ["<stated but unmodeled quantity>"],
  "confidence": <0..1>,
  "notes": null,
  "ambiguity_flags": []
}
- `type:"value"` = fix the feature exactly. `type:"range"` = bound it (uniform/midpoint -> a drawn observed value; reject -> keep free and filter generated samples to the range).
- No time fields and no trajectories are used in this mode.

# EXAMPLE
INPUT: "Generate 200 VV-ECMO records with low oxygen saturation and arterial oxygen pressure."
OUTPUT:
{
  "mode": "single_step",
  "scenario_name": "VV-ECMO with low oxygenation",
  "csn_or_template": null,
  "num_samples": 200,
  "seed": 0,
  "conditioning": [
    {
      "feature": "spo2",
      "type": "range",
      "min": 85,
      "max": 92,
      "sampling": "uniform",
      "units": "%
    },
    {
      "feature": "partial_pressure_of_oxygen_(pao2)",
      "type": "range",
      "min": 50,
      "max": 80,
      "sampling": "uniform",
      "units": "mmHg"
    }
  ],
  "unsupported_facts": [],
  "confidence": 0.9,
  "notes": null,
  "ambiguity_flags": []
}

# USER REQUEST
<<<clinician utterance here>>>
\end{lstlisting}

\begin{lstlisting}[
caption={\revision{Complete Few-Shot Prompt Template for Parsing a Longitudinal Request}},
label={lst:time_series_prompt},
basicstyle=\color{black}\ttfamily\normalsize,
breakatwhitespace=false,
columns=fullflexible
]
# SYSTEM
You convert a clinician's natural-language request (plus any explicit feature values) into ONE JSON object that drives a LONGITUDINAL (time-series) synthetic-trajectory generator over L timesteps. The JSON must validate against `time_series.schema.json`.

RULES (must follow exactly):
- Output ONLY one JSON object, no prose, no markdown fences.
- `feature` MUST be copied verbatim from the catalog below. NEVER invent or abbreviate a feature name.
- Keep numeric values in the listed `unit`, within the `plausible range`, unless the clinician explicitly asks for an extreme.
- Any quantity the clinician mentions that is NOT in the catalog (e.g. FiO2, PEEP, lung compliance, plateau pressure, urine output, ventilator/ECMO settings) goes in `unsupported_facts` as a short string -- NEVER in `conditioning`.
- Put anything you could not map confidently into `ambiguity_flags`; set `confidence` (0..1) to your overall parse confidence.
- Leave any feature you are not told about OUT of `conditioning` (the generator fills it in).
- This is TIME-SERIES mode: include `time_grid` and put time windows on conditions via `window` (step indices OR hours, not both). Use `type:"trajectory"` for time-varying patterns.

# ALLOWED FEATURES (35) -- names, units, plausible ranges
| feature (use this exact name) | unit | plausible range | type |
|---|---|---|---|
| `temperature` | degC | 30 -- 43 | continuous |
| `map_line` | mmHg | 20 -- 160 | continuous |
| `map_cuff` | mmHg | 20 -- 160 | continuous |
| `pulse` | bpm | 20 -- 220 | continuous |
| `unassisted_resp_rate` | breaths/min | 0 -- 60 | continuous |
| `end_tidal_co2` | mmHg | 0 -- 80 | continuous |
| `o2_flow_rate` | L/min | 0 -- 60 | continuous |
| `base_excess` | mmol/L | -30 -- 30 | continuous |
| `bicarb_(hco3)` | mmol/L | 5 -- 50 | continuous |
| `blood_urea_nitrogen_(bun)` | mg/dL | 1 -- 200 | continuous |
| `creatinine` | mg/dL | 0.1 -- 20 | continuous |
| `phosphorus` | mg/dL | 0.5 -- 15 | continuous |
| `hemoglobin` | g/dL | 2 -- 22 | continuous |
| `met_hgb` | %
| `platelets` | 10^3/uL | 1 -- 1000 | continuous |
| `white_blood_cell_count` | 10^3/uL | 0 -- 100 | continuous |
| `carboxy_hgb` | %
| `alanine_aminotransferase_(alt)` | U/L | 0 -- 10000 | continuous |
| `ammonia` | umol/L | 0 -- 500 | continuous |
| `aspartate_aminotransferase_(ast)` | U/L | 0 -- 10000 | continuous |
| `bilirubin_total` | mg/dL | 0 -- 50 | continuous |
| `fibrinogen` | mg/dL | 30 -- 1000 | continuous |
| `inr` | ratio | 0.5 -- 12 | continuous |
| `lactate_dehydrogenase` | U/L | 50 -- 10000 | continuous |
| `lactic_acid` | mmol/L | 0 -- 30 | continuous |
| `partial_prothrombin_time_(ptt)` | sec | 10 -- 200 | continuous |
| `lipase` | U/L | 0 -- 10000 | continuous |
| `partial_pressure_of_carbon_dioxide_(paco2)` | mmHg | 10 -- 150 | continuous |
| `ph` | pH | 6.6 -- 7.8 | continuous |
| `saturation_of_oxygen_(sao2)` | %
| `procalcitonin` | ng/mL | 0 -- 500 | continuous |
| `gcs_total_score` | score | 3 -- 15 | ordinal |
| `spo2` | %
| `partial_pressure_of_oxygen_(pao2)` | mmHg | 20 -- 700 | continuous |
| `rass_score` | score | -5 -- 4 | ordinal |

# OUTPUT FORMAT
{
  "mode": "time_series",
  "scenario_name": "<short label>",
  "csn_or_template": null,
  "num_samples": <int >= 1>,
  "seed": <int or null>,
  "time_grid": {"n_timesteps": 96, "hours_per_step": 1.0},
  "conditioning": [
    {"feature": "<name>", "type": "value", "value": <number>, "units": "<unit>", "window": {"start_step": <int>, "end_step": <int>}},
    {"feature": "<name>", "type": "range", "min": <number>, "max": <number>, "sampling": "uniform|midpoint|reject", "units": "<unit>"},
    {"feature": "<name>", "type": "trajectory", "pattern": "increasing|decreasing|stable|linear|spike|drop", "from": <number>, "to": <number>, "units": "<unit>", "window": {"start_hour": <num>, "duration_hours": <num>}}
  ],
  "unsupported_facts": ["<stated but unmodeled quantity>"],
  "confidence": <0..1>,
  "notes": null,
  "ambiguity_flags": []
}
- `window` omitted => whole trajectory. Same feature may use multiple DISJOINT windows (e.g. stable then declining). Overlapping windows on one feature are invalid.
- `trajectory` direction must be consistent: increasing needs to>from; decreasing to<from; stable uses `from` only; spike/drop go baseline->peak/trough->baseline (window >= 3 steps).

# EXAMPLE
INPUT: "Patient stable for the first day, then SpO2 drops from 96 to 80 over hours 24-48 with rising PaCO2 and tachycardia."
OUTPUT:
{
  "mode": "time_series",
  "scenario_name": "Acute hypoxemic deterioration during hours 24-48",
  "csn_or_template": null,
  "num_samples": 50,
  "seed": 3,
  "time_grid": {
    "n_timesteps": 96,
    "hours_per_step": 1.0
  },
  "conditioning": [
    {
      "feature": "spo2",
      "type": "trajectory",
      "pattern": "stable",
      "from": 96,
      "units": "%
      "window": {
        "start_step": 0,
        "end_step": 24
      }
    },
    {
      "feature": "spo2",
      "type": "trajectory",
      "pattern": "decreasing",
      "from": 96,
      "to": 80,
      "units": "%
      "window": {
        "start_hour": 24,
        "duration_hours": 24
      }
    },
    {
      "feature": "partial_pressure_of_carbon_dioxide_(paco2)",
      "type": "trajectory",
      "pattern": "increasing",
      "from": 45,
      "to": 70,
      "window": {
        "start_hour": 24,
        "duration_hours": 24
      }
    },
    {
      "feature": "pulse",
      "type": "range",
      "min": 100,
      "max": 130,
      "sampling": "uniform",
      "units": "bpm",
      "window": {
        "start_hour": 24,
        "duration_hours": 24
      }
    }
  ],
  "notes": "Two disjoint windows on spo2 (stable 0-24h, declining 24-48h); hour-based windows; steps 48-96 left free.",
  "ambiguity_flags": []
}

# USER REQUEST
<<<clinician utterance here>>>
\end{lstlisting}

\begin{lstlisting}[
caption={\revision{JSON Schema for a Request at a Single Time Point}},
label={lst:single_step_schema},
basicstyle=\color{black}\ttfamily\normalsize,
breakatwhitespace=false,
columns=fullflexible
]
{
  "$schema": "https://json-schema.org/draft/2020-12/schema",
  "$id": "https://ige.local/schema/single_step.schema.json",
  "title": "IGE single-time-point generation request",
  "description": "Constrains the LLM output for SINGLE-TIME-POINT generation (IGE-PMVAE / IGE-VAEAC / IGE-TabDDPM). The LLM reads a clinician's free-text description plus any explicit feature values and emits THIS object. `conditioning` fixes/bounds observed features; everything else is generated. No time dimension is used here (use time_series.schema.json for trajectories).",
  "type": "object",
  "additionalProperties": false,
  "required": [
    "scenario_name",
    "conditioning",
    "num_samples"
  ],
  "properties": {
    "mode": {
      "const": "single_step"
    },
    "scenario_name": {
      "type": "string",
      "minLength": 1
    },
    "csn_or_template": {
      "type": [
        "string",
        "null"
      ],
      "description": "Seed/condition on a real patient CSN or a named template; null = generate from scratch."
    },
    "num_samples": {
      "type": "integer",
      "minimum": 1,
      "default": 100
    },
    "seed": {
      "type": [
        "integer",
        "null"
      ]
    },
    "conditioning": {
      "type": "array",
      "items": {
        "$ref": "#/$defs/condition"
      }
    },
    "unsupported_facts": {
      "type": "array",
      "items": {
        "type": "string"
      },
      "description": "Clinically stated but NOT one of the 35 modeled features (e.g. FiO2, PEEP, lung compliance, plateau pressure, urine output, ventilator/ECMO settings). Recorded for provenance; the model CANNOT generate them. Do NOT put these in `conditioning`."
    },
    "confidence": {
      "type": "number",
      "minimum": 0,
      "maximum": 1
    },
    "notes": {
      "type": [
        "string",
        "null"
      ]
    },
    "ambiguity_flags": {
      "type": "array",
      "items": {
        "type": "string"
      },
      "description": "Phrases the parser could not resolve; non-empty should trigger a clarifying question, not generation."
    }
  },
  "$defs": {
    "feature": {
      "description": "Canonical feature; MUST be one of the 35 model features (feature_dictionary.json).",
      "enum": [
        "temperature",
        "map_line",
        "map_cuff",
        "pulse",
        "unassisted_resp_rate",
        "end_tidal_co2",
        "o2_flow_rate",
        "base_excess",
        "bicarb_(hco3)",
        "blood_urea_nitrogen_(bun)",
        "creatinine",
        "phosphorus",
        "hemoglobin",
        "met_hgb",
        "platelets",
        "white_blood_cell_count",
        "carboxy_hgb",
        "alanine_aminotransferase_(alt)",
        "ammonia",
        "aspartate_aminotransferase_(ast)",
        "bilirubin_total",
        "fibrinogen",
        "inr",
        "lactate_dehydrogenase",
        "lactic_acid",
        "partial_prothrombin_time_(ptt)",
        "lipase",
        "partial_pressure_of_carbon_dioxide_(paco2)",
        "ph",
        "saturation_of_oxygen_(sao2)",
        "procalcitonin",
        "gcs_total_score",
        "spo2",
        "partial_pressure_of_oxygen_(pao2)",
        "rass_score"
      ]
    },
    "condition": {
      "oneOf": [
        {
          "type": "object",
          "additionalProperties": false,
          "description": "Fix the feature to an exact value (observed).",
          "required": [
            "feature",
            "type",
            "value"
          ],
          "properties": {
            "feature": {
              "$ref": "#/$defs/feature"
            },
            "type": {
              "const": "value"
            },
            "value": {
              "type": "number"
            },
            "units": {
              "type": [
                "string",
                "null"
              ]
            }
          }
        },
        {
          "type": "object",
          "additionalProperties": false,
          "description": "Constrain the feature to a range. sampling=uniform|midpoint becomes an observed value; sampling=reject keeps the feature free and filters generated samples to the range.",
          "required": [
            "feature",
            "type",
            "min",
            "max"
          ],
          "properties": {
            "feature": {
              "$ref": "#/$defs/feature"
            },
            "type": {
              "const": "range"
            },
            "min": {
              "type": "number"
            },
            "max": {
              "type": "number"
            },
            "sampling": {
              "enum": [
                "uniform",
                "midpoint",
                "reject"
              ],
              "default": "uniform"
            },
            "units": {
              "type": [
                "string",
                "null"
              ]
            }
          }
        }
      ]
    }
  }
}
\end{lstlisting}

\begin{lstlisting}[
caption={\revision{JSON Schema for a Longitudinal Request}},
label={lst:time_series_schema},
basicstyle=\color{black}\ttfamily\normalsize,
breakatwhitespace=false,
columns=fullflexible
]
{
  "$schema": "https://json-schema.org/draft/2020-12/schema",
  "$id": "https://ige.local/schema/time_series.schema.json",
  "title": "IGE time-series (trajectory) generation request",
  "description": "Constrains the LLM output for LONGITUDINAL generation (IGE-CSDI / IGE-FGTI / IGE-MTSCI). The LLM reads a clinician's free-text description plus any explicit feature values and emits THIS object. `conditioning` fixes/bounds observed (feature, timestep) entries; everything else is generated over the L-step trajectory.",
  "type": "object",
  "additionalProperties": false,
  "required": [
    "scenario_name",
    "time_grid",
    "conditioning",
    "num_samples"
  ],
  "properties": {
    "mode": {
      "const": "time_series"
    },
    "scenario_name": {
      "type": "string",
      "minLength": 1
    },
    "csn_or_template": {
      "type": [
        "string",
        "null"
      ],
      "description": "Seed/condition on a real patient CSN or a named template; null = from scratch."
    },
    "num_samples": {
      "type": "integer",
      "minimum": 1,
      "default": 100
    },
    "seed": {
      "type": [
        "integer",
        "null"
      ]
    },
    "time_grid": {
      "type": "object",
      "additionalProperties": false,
      "required": [
        "n_timesteps"
      ],
      "description": "Trajectory length and hour<->step resolution. Current checkpoints use n_timesteps=96 at 1 h/step.",
      "properties": {
        "n_timesteps": {
          "type": "integer",
          "minimum": 1,
          "default": 96
        },
        "hours_per_step": {
          "type": [
            "number",
            "null"
          ],
          "exclusiveMinimum": 0,
          "default": 1
        }
      }
    },
    "conditioning": {
      "type": "array",
      "items": {
        "$ref": "#/$defs/condition"
      }
    },
    "unsupported_facts": {
      "type": "array",
      "items": {
        "type": "string"
      },
      "description": "Clinically stated but NOT one of the 35 modeled features (FiO2, PEEP, compliance, plateau pressure, urine output, vent/ECMO settings). Recorded only; NOT generatable. Do NOT put in `conditioning`."
    },
    "confidence": {
      "type": "number",
      "minimum": 0,
      "maximum": 1
    },
    "notes": {
      "type": [
        "string",
        "null"
      ]
    },
    "ambiguity_flags": {
      "type": "array",
      "items": {
        "type": "string"
      }
    }
  },
  "$defs": {
    "feature": {
      "description": "Canonical feature; MUST be one of the 35 model features (feature_dictionary.json).",
      "enum": [
        "temperature",
        "map_line",
        "map_cuff",
        "pulse",
        "unassisted_resp_rate",
        "end_tidal_co2",
        "o2_flow_rate",
        "base_excess",
        "bicarb_(hco3)",
        "blood_urea_nitrogen_(bun)",
        "creatinine",
        "phosphorus",
        "hemoglobin",
        "met_hgb",
        "platelets",
        "white_blood_cell_count",
        "carboxy_hgb",
        "alanine_aminotransferase_(alt)",
        "ammonia",
        "aspartate_aminotransferase_(ast)",
        "bilirubin_total",
        "fibrinogen",
        "inr",
        "lactate_dehydrogenase",
        "lactic_acid",
        "partial_prothrombin_time_(ptt)",
        "lipase",
        "partial_pressure_of_carbon_dioxide_(paco2)",
        "ph",
        "saturation_of_oxygen_(sao2)",
        "procalcitonin",
        "gcs_total_score",
        "spo2",
        "partial_pressure_of_oxygen_(pao2)",
        "rass_score"
      ]
    },
    "window": {
      "type": "object",
      "additionalProperties": false,
      "description": "Timestep window the constraint applies to. Provide EITHER step indices OR hours (not both). Omit => whole trajectory. Inclusive start, exclusive end.",
      "properties": {
        "start_step": {
          "type": [
            "integer",
            "null"
          ],
          "minimum": 0
        },
        "end_step": {
          "type": [
            "integer",
            "null"
          ],
          "minimum": 0
        },
        "start_hour": {
          "type": [
            "number",
            "null"
          ],
          "minimum": 0
        },
        "duration_hours": {
          "type": [
            "number",
            "null"
          ],
          "exclusiveMinimum": 0
        }
      }
    },
    "condition": {
      "oneOf": [
        {
          "type": "object",
          "additionalProperties": false,
          "description": "Fix the feature to an exact value over the window.",
          "required": [
            "feature",
            "type",
            "value"
          ],
          "properties": {
            "feature": {
              "$ref": "#/$defs/feature"
            },
            "type": {
              "const": "value"
            },
            "value": {
              "type": "number"
            },
            "units": {
              "type": [
                "string",
                "null"
              ]
            },
            "window": {
              "$ref": "#/$defs/window"
            }
          }
        },
        {
          "type": "object",
          "additionalProperties": false,
          "description": "Constrain the feature to a range over the window. sampling=uniform|midpoint -> observed; reject -> free + post-filter.",
          "required": [
            "feature",
            "type",
            "min",
            "max"
          ],
          "properties": {
            "feature": {
              "$ref": "#/$defs/feature"
            },
            "type": {
              "const": "range"
            },
            "min": {
              "type": "number"
            },
            "max": {
              "type": "number"
            },
            "sampling": {
              "enum": [
                "uniform",
                "midpoint",
                "reject"
              ],
              "default": "uniform"
            },
            "units": {
              "type": [
                "string",
                "null"
              ]
            },
            "window": {
              "$ref": "#/$defs/window"
            }
          }
        },
        {
          "type": "object",
          "additionalProperties": false,
          "description": "A temporal pattern materialized across the window. `from`/`to` are endpoint values (raw units). increasing requires to>from; decreasing to<from; stable uses from only; spike/drop go from baseline to a peak/trough and back.",
          "required": [
            "feature",
            "type",
            "pattern"
          ],
          "properties": {
            "feature": {
              "$ref": "#/$defs/feature"
            },
            "type": {
              "const": "trajectory"
            },
            "pattern": {
              "enum": [
                "increasing",
                "decreasing",
                "stable",
                "linear",
                "spike",
                "drop"
              ]
            },
            "from": {
              "type": [
                "number",
                "null"
              ]
            },
            "to": {
              "type": [
                "number",
                "null"
              ]
            },
            "units": {
              "type": [
                "string",
                "null"
              ]
            },
            "window": {
              "$ref": "#/$defs/window"
            }
          }
        }
      ]
    }
  }
}
\end{lstlisting}

\section{\revision{Supporting Backbone Formulations}}
\label{app:backbone_formulations}

\revision{This appendix provides the standard objectives and sampling equations for the generative backbones summarized in Methods. These formulations follow the cited backbone papers.}

\paragraph{\revision{VAEAC and PMVAE}}

\revision{VAEAC \citep{ivanov2018variational} learns a mask-dependent conditional prior by maximizing}
\begin{align}
    \revision{\mathcal{L}_{\mathrm{VAEAC}}
    =\mathbb{E}_{q_{\phi}(z\mid x,b)}
    [\log p_{\theta}(x_u\mid z,x_o,b)]
    -D_{\mathrm{KL}}\!\left(
    q_{\phi}(z\mid x,b)\,\|\,p_{\psi}(z\mid x_o,b)
    \right).}
\end{align}
\revision{PMVAE \citep{strauss2022posterior} trains a partial encoder by matching the posterior inferred from observed features to the full-data posterior, using}
\begin{align}
    \revision{\mathcal{L}_{\mathrm{PM}}
    =-\mathbb{E}_{z\sim q_{\psi}(\cdot\mid x)}
    [\log q_{\theta}(z\mid x_o)].}
\end{align}

\paragraph{\revision{IGE-TabDDPM}}

\revision{IGE-TabDDPM uses replacement-based conditional sampling \citep{lugmayr2022repaint}. During the reverse update from diffusion step $\tau$ to $\tau-1$, the sampler combines the observed entries at the corresponding noise level, $x_{\mathrm{known}}^{\tau-1}$, with the entries proposed by the denoiser, $x_{\mathrm{sample}}^{\tau-1}$:}
\begin{align}
    \revision{x^{\tau-1}
    =b\odot x_{\mathrm{known}}^{\tau-1}
    +(1-b)\odot x_{\mathrm{sample}}^{\tau-1}.}
\end{align}
\revision{Thus, the observed entries are restored at each reverse-diffusion step while the remaining entries are generated.}

\paragraph{\revision{CSDI}}

\revision{At diffusion step $\tau$, CSDI \citep{tashiro2021csdi} updates the unobserved trajectory according to}
\begin{align}
    \revision{X_u^{\tau-1}
    \leftarrow \frac{1}{\sqrt{\alpha_\tau}}
    \left(
    X_u^\tau
    -\frac{1-\alpha_\tau}{\sqrt{1-\bar{\alpha}_\tau}}
    \epsilon_\theta(X_u^\tau,\tau\mid X_o,B)
    \right)
    +\sigma_\tau z,}
\end{align}
\revision{where $z\sim\mathcal{N}(0,I)$. Its conditional noise-matching objective is}
\begin{align}
    \revision{\mathcal{L}_{\mathrm{CSDI}}
    =\mathbb{E}_{X_o,X_u,\epsilon,\tau}
    \left[
    \|\epsilon-\epsilon_\theta(X_u^\tau,\tau\mid X_o,B)\|^2
    \right].}
\end{align}

\paragraph{\revision{FGTI}}

\revision{FGTI \citep{yang2024frequencyaware} augments conditional noise matching with a frequency-domain term:}
\begin{align}
    \revision{\mathcal{L}_{\mathrm{FGTI}}}
    &\revision{=\mathbb{E}_{X_o,X_u,\epsilon,\tau}
    \left[
    \|\epsilon-\epsilon_\theta(X_u^\tau,\tau\mid X_o,B)\|^2 \right.}
    \notag\\
    &\revision{\left.
    +\lambda\|
    \mathcal{F}(\epsilon)
    -\mathcal{F}(\epsilon_\theta(X_u^\tau,\tau\mid X_o,B))
    \|^2
    \right],}
\end{align}
\revision{where $\mathcal{F}$ denotes the Fourier transform.}

\paragraph{\revision{MTSCI}}

\revision{MTSCI \citep{zhou2024mtsci} mixes the noisy target with an adjacent context window before denoising:}
\begin{align}
    \revision{X_{\mathrm{in}}
    =B\odot X_o
    +(1-B)\odot
    \left(\gamma X_{\mathrm{ctx}}+(1-\gamma)X_u^\tau\right),}
\end{align}
\revision{where $\gamma\in[0,1]$ is sampled from a Beta distribution.}

\section{\revision{Clinical Variables}}
\label{app:clinical_variables}

\revision{The pooled Duke--Emory representation comprises the 35 clinical variables listed in Table~\ref{tab:dataset_variables}. These variables form the input at a single time point and at each time point in a longitudinal trajectory.}

\begin{table*}[!htbp]
\centering
\caption{\revision{Clinical Variables in the Pooled Duke--Emory Representation}}
\label{tab:dataset_variables}
\scriptsize
\begin{tabular}{@{}p{5.25cm}p{1.60cm}p{5.25cm}p{1.60cm}@{}}
\toprule
\revision{Variable} & \revision{Unit} & \revision{Variable} & \revision{Unit} \\
\midrule
\revision{Alanine Aminotransferase (ALT)} & \revision{U/L} &
\revision{Mean Arterial Pressure (Arterial Line)} & \revision{mmHg} \\
\revision{Ammonia} & \revision{$\mu$mol/L} &
\revision{Methemoglobin} & \revision{\%} \\
\revision{Aspartate Aminotransferase (AST)} & \revision{U/L} &
\revision{O$_2$ Flow Rate} & \revision{L/min} \\
\revision{Base Excess} & \revision{mmol/L} &
\revision{Partial Pressure of Carbon Dioxide (PaCO$_2$)} & \revision{mmHg} \\
\revision{Total Bilirubin} & \revision{mg/dL} &
\revision{Partial Pressure of Oxygen (PaO$_2$)} & \revision{mmHg} \\
\revision{Bicarbonate (HCO$_3^-$)} & \revision{mmol/L} &
\revision{Procalcitonin} & \revision{ng/mL} \\
\revision{Blood Urea Nitrogen (BUN)} & \revision{mg/dL} &
\revision{pH} & \revision{Dimensionless} \\
\revision{Carboxyhemoglobin} & \revision{\%} &
\revision{Phosphorus} & \revision{mg/dL} \\
\revision{Creatinine} & \revision{mg/dL} &
\revision{Platelet Count} & \revision{$10^3$/$\mu$L} \\
\revision{End-Tidal CO$_2$} & \revision{mmHg} &
\revision{Partial Thromboplastin Time (PTT)} & \revision{sec} \\
\revision{Fibrinogen} & \revision{mg/dL} &
\revision{Pulse} & \revision{bpm} \\
\revision{Glasgow Coma Scale (GCS) Total Score} & \revision{Score} &
\revision{Richmond Agitation--Sedation Scale (RASS) Score} & \revision{Score} \\
\revision{Hemoglobin} & \revision{g/dL} &
\revision{Respiratory Rate (Unassisted)} & \revision{breaths/min} \\
\revision{International Normalized Ratio (INR)} & \revision{Dimensionless} &
\revision{Arterial Oxygen Saturation (SaO$_2$)} & \revision{\%} \\
\revision{Lactic Acid} & \revision{mmol/L} &
\revision{Peripheral Oxygen Saturation (SpO$_2$)} & \revision{\%} \\
\revision{Lactate Dehydrogenase} & \revision{U/L} &
\revision{Temperature} & \revision{$^\circ$C} \\
\revision{Lipase} & \revision{U/L} &
\revision{White Blood Cell Count} & \revision{$10^3$/$\mu$L} \\
\revision{Mean Arterial Pressure (Cuff)} & \revision{mmHg} &
& \\
\bottomrule
\end{tabular}
\end{table*}

\section{Additional Experiments}

\subsection{LLM Module Evaluation}
\label{sec:llm_eval}
We evaluate the optional LLM parser, which converts natural-language clinical requests into structured JSON specifications. We construct evaluation datasets to assess Gemma3-4B, MedGemma-4B, \revision{MedAlpaca 7B,} Mistral-7B, and Qwen3-4B \citep{gemmateam2025gemma3technicalreport, sellergren2025medgemmatechnicalreport, jiang2023mistral7b, yang2025qwen3technicalreport}.
All prompt families share a common system instruction:
\begin{lstlisting}
"Your task is to convert clinician requests into structured JSON specifications for generating synthetic clinical data. Be precise, thorough, and always include a confidence score."
\end{lstlisting}
The only exception is the role-anchored variant, where the system role is
specified as: ``You are an expert clinical data specification assistant.'' The differences across prompt families lie primarily in the user prompts. We illustrate these differences using the following clinician utterance:
\textit{``Severe ARDS patient on VV-ECMO with gradual improvement in oxygenation over 48 hours.''}

Below are the prompt templates for each variant, all applied to the same clinician utterance.
\begin{enumerate}
\item Direct Zero-Shot
\begin{lstlisting}
Convert the following clinician utterances into a JSON object that follows this schema:
<JSON_SCHEMA>
Here is a variable dictionary for mapping clinical terms to feature names:
<VARIABLE_DICTIONARY>
<CLINICIAN_UTTERANCE>
\end{lstlisting}

\item Few-Shot
\begin{lstlisting}
Convert clinician utterance into structured JSON. The JSON must follow this schema:
<JSON_SCHEMA>
Here is a variable dictionary for mapping clinical terms to feature names:
<VARIABLE_DICTIONARY>
Study the following examples carefully:
<few-shot examples>
Now convert this utterance using the same schema and variable mappings:
<CLINICIAN_UTTERANCE>
\end{lstlisting}

\item Role-Anchored Few-Shot
\begin{lstlisting}
You are an expert clinical data specification assistant. Here are examples of how to convert clinician utterances into JSON. The JSON must conform to this schema:
<JSON_SCHEMA>
Here is a variable dictionary for mapping clinical terms to feature names:
<VARIABLE_DICTIONARY>
Study these examples:
<few-shot examples>
Now convert the following clinician utterance. Include a confidence score.
<CLINICIAN_UTTERANCE>
\end{lstlisting}

\item Stepwise Few-Shot
\begin{lstlisting}
Follow these examples to learn the task:
<few-shot examples>
For the next utterance, follow these steps systematically:
Step 1: Extract the scenario name and time window.
Step 2: Map clinical parameters to feature names.
Step 3: Determine type (fixed value, range, or trajectory).
Step 4: Construct the JSON following this schema:
<JSON_SCHEMA>
Here is a variable dictionary for mapping clinical terms to feature names:
<VARIABLE_DICTIONARY>
Step 5: Add notes about ambiguities.
<CLINICIAN_UTTERANCE>
\end{lstlisting}

\end{enumerate}

\Cref{tab:llm_results} summarizes the performance of the \revision{optional LLM parser} across four prompt formats and \revision{five} large language models. We report syntactic metrics (parse success rate and schema compliance), semantic metrics (slot-level F1 and \revision{spurious slot rate}), and the ambiguity flagging rate.

\begin{table}[!t]
\centering
\caption{LLM Module Performance}
\label{tab:llm_results}
\renewcommand{\arraystretch}{1.15}
\setlength{\tabcolsep}{3pt}
\begin{tabular}{@{}llccccc@{}}
\toprule
\textbf{Prompt} & \textbf{Model} & \textbf{Parse} & \textbf{Schema} & \textbf{Slot} & \revision{\textbf{Spurious}} & \textbf{Ambig.} \\
 & & \textbf{(\%)}  & \textbf{(\%)} & \textbf{F1} & \textbf{(\%)} & \textbf{(\%)} \\
\midrule
\multirow{5}{*}{\shortstack[l]{\revision{Direct}\\Zero-Shot}}
 & Gemma3 4B & 100 & 100 & 0.74 & 5.2 & 0.0 \\
 & MedGemma 4B & 100 & 96.7 & 0.60 & 10.3 & 0.0 \\
 & \revision{MedAlpaca 7B} & \revision{85.0} & \revision{47.5} & \revision{0.23} & \revision{17.3} & \revision{0.0} \\
 & Mistral 7B & 91.7 & 0.0 & 0.00 & 0.0 & 0.0 \\
 & Qwen3 4B & 100 & 100 & 0.77 & 13.5 & 5.0 \\
\midrule
\multirow{5}{*}{\shortstack[l]{Few-Shot}}
 & Gemma3 4B & 100 & 100 & \textbf{0.97} & \textbf{2.7} & 8.3 \\
 & MedGemma 4B & 100 & 97.5 & 0.93 & 3.3 & 10.8 \\
 & \revision{MedAlpaca 7B} & \revision{45.8} & \revision{41.7} & \revision{0.32} & \revision{2.0} & \revision{6.7} \\
 & Mistral 7B & 73.3 & 71.7 & 0.67 & 4.3 & 18.3 \\
 & Qwen3 4B & 100 & 97.5 & 0.90 & 3.9 & 5.0 \\
\midrule
\multirow{5}{*}{\shortstack[l]{Role-Anchored\\Few-Shot}}
 & Gemma3 4B & 100 & 98.3 & 0.93 & 4.1 & 9.2 \\
 & MedGemma 4B & 100 & 99.2 & 0.94 & 3.4 & 12.5 \\
 & \revision{MedAlpaca 7B} & \revision{40.0} & \revision{35.8} & \revision{0.30} & \revision{0.7} & \revision{0.0} \\
 & Mistral 7B & 99.2 & 99.2 & 0.92 & 4.5 & 19.2 \\
 & Qwen3 4B & 100 & 100 & 0.90 & 5.4 & 5.0 \\
\midrule
\multirow{5}{*}{\shortstack[l]{Stepwise\\Few-Shot}}
 & Gemma3 4B & 100 & 100 & 0.94 & 4.7 & 19.2 \\
 & MedGemma 4B & 98.3 & 95.8 & 0.87 & 6.2 & 19.2 \\
 & \revision{MedAlpaca 7B} & \revision{0.0} & \revision{0.0} & \revision{0.00} & \revision{0.0} & \revision{0.0} \\
 & Mistral 7B & 99.2 & 96.7 & 0.89 & 5.7 & 20.0 \\
 & Qwen3 4B & 100 & 100 & 0.94 & 4.2 & 7.5 \\
\bottomrule
\end{tabular}
\end{table}

\revision{Gemma3 4B with Few-Shot prompting achieves the best overall parsing performance, with a slot F1 of 0.97, 100\% parse success, 100\% schema compliance, and a 2.7\% spurious slot rate. Few-Shot prompting gives a higher slot F1 than Direct Zero-Shot prompting for all five models. Biomedical instruction tuning does not consistently improve parsing: MedGemma 4B reaches a slot F1 of 0.94 with Role-Anchored Few-Shot prompting, whereas MedAlpaca 7B reaches at most 0.32 and often returns output that cannot be parsed or does not follow the schema. MedAlpaca's 0\% spurious slot rate under Stepwise Few-Shot prompting results from complete parsing failure rather than correct extraction.}

\subsection{Subset Evaluation on Clinically Salient Features} 
\label{subsec:subset_evaluation}

We evaluate model performance when conditioning on a subset of clinically salient features identified in consultation with domain experts.
Specifically, we collaborated with clinicians to select 12 key variables critical to the management of VV-ECMO 
patients in our dataset.
Focusing on this targeted subset allows us to simulate real-world clinical scenarios where practitioners prioritize specific physiological indicators. 
These features serve as the observed conditioning inputs $x_o$ to guide the generation of the remaining patient features.
The selected features are grouped into three primary physiological categories:

\begin{enumerate}
    \item \textbf{Oxygenation-related features:} Indicators of systemic oxygenation, including peripheral oxygen saturation ($\text{SpO}_2$), partial pressure of oxygen ($\text{PaO}_2$), and arterial oxygen saturation ($\text{SaO}_2$).
    \item \textbf{Ventilation and acid-base features:} Indicators of ventilation and acid-base balance, including base excess, partial pressure of carbon dioxide ($\text{PaCO}_2$), bicarbonate ($\text{HCO}_3^-$), and pH.
    \item \textbf{Laboratory and organ function markers:} Laboratory values related to hemolysis, renal function, and hepatic function, including hemoglobin, platelet count, creatinine, lactate dehydrogenase, and total bilirubin.
\end{enumerate}

\revision{To evaluate the framework on a targeted data-imputation task, we compare the InterGenEHR$^+$ variants with VAE, EHRDiff, and TabDiff.}
We keep these 12 clinically salient features unmasked as the conditioning inputs $x_o$ and mask the remaining variables as the generation targets $x_u$.

\revision{The results are summarized in \Cref{tab:Subset}.}
\revision{IGE-VAEAC achieves the lowest MSE of $0.3058$, while TabDiff achieves the lowest NLL of $0.7166$. IGE-PMVAE also performs well, with an MSE of $0.4566$ and NLL of $0.7651$.}
\revision{These results show that the framework can generate the unobserved features in a patient record from a clinician-selected subset of observed features.}

\begin{table}[ht]
\centering
\caption{Imputation with a Clinically Salient Conditioning Subset. \algnameAbbrv\ denotes generative backbones integrated into the \algname{} framework.}
\label{tab:Subset}
\begin{tabular}{lcc}
\toprule
Method & MSE $\downarrow$ & NLL $\downarrow$ \\ 
\midrule
VAE          & 1.4482   & 4.6786    \\
\revision{EHRDiff} & \revision{1.5661} & \revision{0.7830} \\
\revision{TabDiff} & \revision{1.4331} & \revision{0.7166} \\
\algnameAbbrv-PMVAE   & 0.4566   & 0.7651    \\
\algnameAbbrv-VAEAC   & 0.3058   & 1.2141    \\
\algnameAbbrv-TabDDPM & 0.8091   & 0.8330    \\
\bottomrule
\end{tabular}
\end{table}

\subsection{More Results on In-Hospital Mortality Prediction}
\label{sec:more_downstream_results}

\revision{\Cref{tab:classification_results_lr_gbdt} provides additional results for \revisiontwo{LR, GBDT, and MLP}. These results compare classification performance when training on synthetic data with the corresponding real-data reference.}

\begin{table*}[!t]
\centering
\caption{\revision{Additional Downstream Utility Results for In-Hospital Mortality Classification Using \revisiontwo{Logistic Regression, Gradient-Boosted Decision Trees, and a Multilayer Perceptron} (Mean $\pm$ Standard Deviation Across Five Independent Seeds)}}
\label{tab:classification_results_lr_gbdt}
\resizebox{\textwidth}{!}{
\begin{tabular}{llcccccc}
\toprule
\revision{Classifier}
& \revision{Training data}
& \revision{AUROC $\uparrow$}
& \revision{AUPRC $\uparrow$}
& \revision{Accuracy $\uparrow$}
& \revision{Precision $\uparrow$}
& \revision{Recall $\uparrow$}
& \revision{F1 $\uparrow$} \\
\midrule

\revision{LR}
& \revision{Real}
& \revision{$0.850 \pm 0.025$}
& \revision{$0.823 \pm 0.025$}
& \revision{$0.839 \pm 0.018$}
& \revision{$0.805 \pm 0.034$}
& \revision{$0.654 \pm 0.046$}
& \revision{$0.714 \pm 0.038$} \\

\revision{LR}
& \revision{\algnameAbbrv-PMVAE}
& \revision{$0.891 \pm 0.068$}
& \revision{$0.885 \pm 0.066$}
& \revision{$0.847 \pm 0.060$}
& \revision{$0.830 \pm 0.114$}
& \revision{$0.829 \pm 0.140$}
& \revision{$0.814 \pm 0.075$} \\

\revision{LR}
& \revision{\algnameAbbrv-VAEAC}
& \revision{$0.849 \pm 0.051$}
& \revision{$0.801 \pm 0.089$}
& \revision{$0.812 \pm 0.044$}
& \revision{$0.774 \pm 0.060$}
& \revision{$0.771 \pm 0.070$}
& \revision{$0.771 \pm 0.054$} \\

\revision{LR}
& \revision{\algnameAbbrv-TabDDPM}
& \revision{$0.849 \pm 0.064$}
& \revision{$0.832 \pm 0.063$}
& \revision{$0.776 \pm 0.069$}
& \revision{$0.748 \pm 0.063$}
& \revision{$0.686 \pm 0.167$}
& \revision{$0.706 \pm 0.115$} \\

\revision{LR}
& \revision{EHRDiff}
& \revision{$0.877 \pm 0.055$}
& \revision{$0.873 \pm 0.045$}
& \revision{$0.835 \pm 0.044$}
& \revision{$0.826 \pm 0.110$}
& \revision{$0.800 \pm 0.146$}
& \revision{$0.796 \pm 0.063$} \\

\revision{LR}
& \revision{TabDiff}
& \revision{$0.820 \pm 0.041$}
& \revision{$0.788 \pm 0.062$}
& \revision{$0.765 \pm 0.074$}
& \revision{$0.719 \pm 0.112$}
& \revision{$0.743 \pm 0.057$}
& \revision{$0.726 \pm 0.068$} \\

\revision{LR}
& \revision{VAE}
& \revision{$0.854 \pm 0.072$}
& \revision{$0.793 \pm 0.110$}
& \revision{$0.812 \pm 0.044$}
& \revision{$0.754 \pm 0.076$}
& \revision{$0.829 \pm 0.107$}
& \revision{$0.783 \pm 0.051$} \\

\midrule

\revision{GBDT}
& \revision{Real}
& \revision{$1.000 \pm 0.000$}
& \revision{$1.000 \pm 0.000$}
& \revision{$0.906 \pm 0.029$}
& \revision{$0.817 \pm 0.048$}
& \revision{$1.000 \pm 0.000$}
& \revision{$0.898 \pm 0.029$} \\

\revision{GBDT}
& \revision{\algnameAbbrv-PMVAE}
& \revision{$0.946 \pm 0.028$}
& \revision{$0.928 \pm 0.038$}
& \revision{$0.824 \pm 0.053$}
& \revision{$0.870 \pm 0.113$}
& \revision{$0.686 \pm 0.057$}
& \revision{$0.763 \pm 0.066$} \\

\revision{GBDT}
& \revision{\algnameAbbrv-VAEAC}
& \revision{$0.943 \pm 0.039$}
& \revision{$0.940 \pm 0.039$}
& \revision{$0.882 \pm 0.074$}
& \revision{$0.860 \pm 0.085$}
& \revision{$0.857 \pm 0.156$}
& \revision{$0.851 \pm 0.105$} \\

\revision{GBDT}
& \revision{\algnameAbbrv-TabDDPM}
& \revision{$0.929 \pm 0.059$}
& \revision{$0.893 \pm 0.114$}
& \revision{$0.847 \pm 0.088$}
& \revision{$0.870 \pm 0.113$}
& \revision{$0.743 \pm 0.140$}
& \revision{$0.797 \pm 0.116$} \\

\revision{GBDT}
& \revision{EHRDiff}
& \revision{$0.897 \pm 0.070$}
& \revision{$0.881 \pm 0.044$}
& \revision{$0.800 \pm 0.029$}
& \revision{$0.820 \pm 0.016$}
& \revision{$0.657 \pm 0.070$}
& \revision{$0.728 \pm 0.050$} \\

\revision{GBDT}
& \revision{TabDiff}
& \revision{$0.906 \pm 0.066$}
& \revision{$0.896 \pm 0.053$}
& \revision{$0.859 \pm 0.047$}
& \revision{$0.880 \pm 0.062$}
& \revision{$0.771 \pm 0.146$}
& \revision{$0.811 \pm 0.074$} \\

\revision{GBDT}
& \revision{VAE}
& \revision{$0.874 \pm 0.071$}
& \revision{$0.845 \pm 0.092$}
& \revision{$0.776 \pm 0.101$}
& \revision{$0.843 \pm 0.139$}
& \revision{$0.571 \pm 0.239$}
& \revision{$0.650 \pm 0.199$} \\

\midrule

\revision{\revisiontwo{MLP}}
& \revision{Real} &
\revision{$0.943 \pm 0.018$} &
\revision{$0.903 \pm 0.044$} &
\revision{$\boldsymbol{0.929 \pm 0.024}$} &
\revision{$0.871 \pm 0.007$} &
\revision{$\boldsymbol{0.971 \pm 0.057}$} &
\revision{$\boldsymbol{0.918 \pm 0.030}$} \\

\revision{\revisiontwo{MLP}}
& \revision{IGE-PMVAE} &
\revision{$0.940 \pm 0.028$} &
\revision{$0.884 \pm 0.080$} &
\revision{$0.882 \pm 0.037$} &
\revision{$0.856 \pm 0.013$} &
\revision{$0.857 \pm 0.090$} &
\revision{$0.855 \pm 0.052$} \\

\revision{\revisiontwo{MLP}}
& \revision{IGE-VAEAC} &
\revision{$0.926 \pm 0.025$} &
\revision{$0.905 \pm 0.036$} &
\revision{$0.847 \pm 0.047$} &
\revision{$0.798 \pm 0.078$} &
\revision{$0.857 \pm 0.000$} &
\revision{$0.824 \pm 0.043$} \\

\revision{\revisiontwo{MLP}}
& \revision{IGE-TabDDPM} &
\revision{$0.906 \pm 0.047$} &
\revision{$0.897 \pm 0.056$} &
\revision{$0.859 \pm 0.060$} &
\revision{$0.818 \pm 0.077$} &
\revision{$0.857 \pm 0.090$} &
\revision{$0.833 \pm 0.067$} \\

\revision{\revisiontwo{MLP}}
& \revision{EHRDiff} &
\revision{$\boldsymbol{0.946 \pm 0.046}$} &
\revision{$\boldsymbol{0.952 \pm 0.034}$} &
\revision{$0.906 \pm 0.029$} &
\revision{$\boldsymbol{0.898 \pm 0.088}$} &
\revision{$0.886 \pm 0.057$} &
\revision{$0.887 \pm 0.030$} \\

\revision{\revisiontwo{MLP}}
& \revision{TabDiff} &
\revision{$0.869 \pm 0.063$} &
\revision{$0.847 \pm 0.059$} &
\revision{$0.847 \pm 0.071$} &
\revision{$0.804 \pm 0.073$} &
\revision{$0.829 \pm 0.107$} &
\revision{$0.815 \pm 0.087$} \\

\revision{\revisiontwo{MLP}}
& \revision{VAE} &
\revision{$0.883 \pm 0.052$} &
\revision{$0.797 \pm 0.085$} &
\revision{$0.847 \pm 0.047$} &
\revision{$0.808 \pm 0.055$} &
\revision{$0.829 \pm 0.107$} &
\revision{$0.815 \pm 0.062$} \\

\bottomrule
\end{tabular}
}
\end{table*}

\section{\revision{Clinician Evaluation Protocol}}
\label{app:clinician_evaluation}

\revisiontwo{The clinician evaluation uses IGE-FGTI trained on the combined Duke--Emory dataset with the same 35 clinical variables as the main experiments (Table~\ref{tab:dataset_variables}).}

\paragraph{\revision{Evaluation procedure}}

\revision{Critical-care clinicians review each requested scenario together with its generated trajectories. The trajectory order is randomized, and the generating model is not disclosed. For each trajectory, the clinicians rate clinical plausibility, internal physiological consistency, requested-scenario representation, physiological possibility, and \revisiontwo{agreement with the requested scenario}. All questions use a 1--5 scale.}

\paragraph{\revision{Survey instrument}}

\revision{Table~\ref{tab:clinician_survey} provides the exact five questions and response scales shown to the clinicians. Q1, Q2, and Q4 assess the clinical realism of the displayed trajectory, while Q3 and Q5 assess its agreement with the requested scenario.}

\begin{table}[!htbp]
\centering
\caption{\revision{Exact Questions and Response Scales Used in the Clinician Evaluation}}
\label{tab:clinician_survey}
\footnotesize
\begin{tabular}{@{}p{0.65cm}p{11.0cm}p{5.0cm}@{}}
\toprule
\revision{Item} & \revision{Evaluation question} & \revision{Response} \\
\midrule
\revision{Q1} &
\revisiontwo{Does this trajectory represent a clinically plausible VV-ECMO patient?} &
\revision{1 (completely implausible) to 5 (completely plausible)} \\
\revision{Q2} &
\revisiontwo{Are the physiological trends internally consistent?} &
\revision{1 (completely inconsistent) to 5 (completely consistent)} \\
\revision{Q3} &
\revisiontwo{Would you consider this trajectory representative of the requested scenario?} &
\revision{1 (completely unrepresentative) to 5 (completely representative)} \\
\revision{Q4} &
\revisiontwo{All values appear physiologically possible?} &
\revision{1 (completely impossible) to 5 (completely possible)} \\
\revision{Q5} &
\revisiontwo{How well does the generated trajectory match this requested scenario?} &
\revision{1 (does not match at all) to 5 (perfectly matches)} \\
\bottomrule
\end{tabular}
\end{table}

\paragraph{\revision{Scenario definitions}}

\revision{The 10 requests in Table~\ref{tab:clinician_scenarios} cover \revisiontwo{severe lung injury, low blood oxygen levels, respiratory and metabolic acidosis, recovery and possible ECMO removal, hemolysis, impaired carbon dioxide clearance, acute kidney injury, and thrombocytopenia} during VV-ECMO support. The original request is shown with each generated trajectory so that clinical plausibility and agreement with the requested scenario can be evaluated separately.}

\begin{table}[!htbp]
\centering
\caption{\revision{Clinical Scenario Requests Used in the Clinician Evaluation}}
\label{tab:clinician_scenarios}
\footnotesize
\begin{tabular}{@{}p{0.55cm}p{3.5cm}p{12.7cm}@{}}
\toprule
\revision{ID} & \revision{Scenario} & \revision{Clinician request} \\
\midrule
\revision{1} &
\revisiontwo{Severe ARDS} &
\revisiontwo{Generate a 48-hour synthetic ICU trajectory for an adult with severe ARDS on VV-ECMO. Include ventilation, ECMO support, oxygenation, hemodynamics, blood gases, and routine labs. Maintain physiologic consistency; use realistic intermittent lab timing and compatible arterial-line and cuff-pressure measurements.} \\
\revision{2} &
\revisiontwo{Refractory hypoxemic respiratory failure} &
\revisiontwo{Generate a 72-hour synthetic ICU trajectory for an adult on VV-ECMO with refractory hypoxemic respiratory failure. Include ECMO, ventilation, hemodynamics, blood gases, and routine labs. Model a clinically coherent course without forcing simple linear trends or incompatible measurements.} \\
\revision{3} &
\revisiontwo{Respiratory acidosis} &
\revisiontwo{Generate a 24-hour synthetic ICU trajectory for an adult on VV-ECMO with respiratory acidosis. Include ECMO and ventilator management, blood gases, hemodynamics, and relevant labs. Ensure plausible relationships among variables, intermittent lab collection, and physiologically possible values.} \\
\revision{4} &
\revisiontwo{Metabolic acidosis} &
\revisiontwo{Generate a 72-hour synthetic ICU trajectory for an adult on VV-ECMO with metabolic acidosis during critical illness. Include hemodynamics, blood gases, ECMO and ventilator support, and relevant labs. Avoid impossible values and unexplained shifts between severe dysfunction and normal results.} \\
\revision{5} &
\revisiontwo{Recovery and possible ECMO removal} &
\revisiontwo{Generate a 72-hour synthetic ICU trajectory for an adult on VV-ECMO being evaluated for recovery and possible liberation from ECMO. Include ventilation, ECMO settings, oxygenation, hemodynamics, and routine labs. Model a nuanced, physiologically plausible course without assuming uniform improvement.} \\
\revision{6} &
\revisiontwo{Suspected hemolysis} &
\revisiontwo{Generate a 72-hour synthetic ICU trajectory for an adult on VV-ECMO with suspected hemolysis. Include ECMO and ventilator support, respiratory and hemodynamic monitoring, blood gases, and relevant hematologic and chemistry labs. Create a plausible pattern without forcing linear laboratory trends.} \\
\revision{7} &
\revisiontwo{Acute oxygenation deterioration} &
\revisiontwo{Generate a 12-hour synthetic ICU trajectory for an adult on VV-ECMO with acute worsening oxygenation. Include ECMO support, ventilation, blood gases, hemodynamics, and relevant labs. Maintain coherence among respiratory variables and avoid arbitrary linear trends.} \\
\revision{8} &
\revisiontwo{Impaired carbon dioxide clearance} &
\revisiontwo{Generate a 48-hour synthetic ICU trajectory for an adult on VV-ECMO with impaired carbon dioxide clearance. Include ECMO settings, ventilation, blood gases, acid-base status, hemodynamics, and routine labs. Use plausible variable relationships and intermittent lab sampling.} \\
\revision{9} &
\revisiontwo{Acute kidney injury} &
\revisiontwo{Generate a 72-hour synthetic ICU trajectory for an adult on VV-ECMO who develops acute kidney injury. Include hemodynamics, fluid-related context, blood gases, ECMO and ventilator support, and renal and metabolic labs. Avoid implausible fluctuations between severe injury and normal renal function.} \\
\revision{10} &
\revisiontwo{Progressive thrombocytopenia} &
\revisiontwo{Generate a 72-hour synthetic ICU trajectory for an adult on VV-ECMO with progressive thrombocytopenia and concern for an ECMO-related hematologic complication. Include ECMO support, ventilation, hemodynamics, blood gases, and hematologic and coagulation labs. Use realistic sampling and avoid forced linear trends.} \\
\bottomrule
\end{tabular}
\end{table}

\paragraph{\revision{LLM parsing of the scenario requests}}

\revision{To examine how the optional LLM parser processes the same 10 scenario requests, Tables~\ref{tab:clinician_parsed_specs_1} and~\ref{tab:clinician_parsed_specs_2} summarize the longitudinal JSON specifications produced before clinician review. This parsing analysis is reported separately from the clinician ratings. Each specification uses a 96-step grid with one-hour time points, and all 10 outputs are parseable and follow the JSON schema. Schema compliance verifies the JSON structure but does not establish whether the extracted values, ranges, or directions are clinically appropriate. Because the saved outputs do not record later clinician confirmation or modification, the tables describe the parser outputs rather than the final clinician-confirmed constraints.}

\begin{table}[!htbp]
\centering
\caption{\revision{LLM-Parsed Longitudinal Specifications Before Clinician Review: Scenarios 1--5}}
\label{tab:clinician_parsed_specs_1}
\scriptsize
\renewcommand{\arraystretch}{1.10}
\begin{tabular}{@{}p{0.55cm}p{4.1cm}p{12.0cm}@{}}
\toprule
\revision{ID} & \revision{Parsed scenario name} & \revision{Parsed conditioning description} \\
\midrule
\revision{1} &
\revision{Gradual ECMO-assisted oxygenation} &
\revision{SaO$_2$: $60\%\rightarrow95\%$ over 0--48 h.} \\
\revision{2} &
\revision{Refractory hypoxemia with ECMO} &
\revision{SaO$_2$: $85\%\rightarrow70\%$ over 0--96 h; arterial-line MAP 50--80 mmHg and cuff MAP 40--70 mmHg over 0--24 h; pulse 120 bpm; end-tidal CO$_2$ 40--60 mmHg.} \\
\revision{3} &
\revision{Worsening respiratory acidosis} &
\revision{PaCO$_2$: $50\rightarrow80$ mmHg and pH: $7.35\rightarrow7.20$ over 0--24 h.} \\
\revision{4} &
\revision{Progressive metabolic acidosis with stable oxygenation} &
\revision{Base excess: $-30\rightarrow-80$ mmol/L and lactic acid: $1\rightarrow8$ mmol/L over 0--96 h. Both are labeled as increasing in the saved JSON.} \\
\revision{5} &
\revision{ECMO recovery and weaning} &
\revision{SaO$_2$: $70\%\rightarrow95\%$ over 0--72 h; arterial-line MAP 25--40 mmHg over 0--24 h; cuff MAP 25--40 mmHg over 24--48 h; O$_2$ flow: $15\rightarrow0$ L/min over 24--72 h; arterial-line MAP: $40\rightarrow20$ mmHg over 72--96 h.} \\
\bottomrule
\end{tabular}
\end{table}

\begin{table}[!htbp]
\centering
\caption{\revision{LLM-Parsed Longitudinal Specifications Before Clinician Review: Scenarios 6--10}}
\label{tab:clinician_parsed_specs_2}
\scriptsize
\renewcommand{\arraystretch}{1.10}
\begin{tabular}{@{}p{0.55cm}p{4.1cm}p{12.0cm}@{}}
\toprule
\revision{ID} & \revision{Parsed scenario name} & \revision{Parsed conditioning description} \\
\midrule
\revision{6} &
\revision{Hemolysis with increasing lactate dehydrogenase and decreasing hemoglobin} &
\revision{Lactate dehydrogenase: $100\rightarrow500$ U/L and hemoglobin: $12\rightarrow8$ g/dL over 0--96 h.} \\
\revision{7} &
\revision{ECMO deterioration} &
\revision{O$_2$ flow: $20\rightarrow40$ L/min, arterial-line MAP 40--80 mmHg, and unassisted respiratory rate 10--30 breaths/min over 0--12 h.} \\
\revision{8} &
\revision{ECMO improvement over 48 hours} &
\revision{End-tidal CO$_2$: $50\rightarrow40$ mmHg over 0--48 h, labeled as increasing in the saved JSON; arterial-line MAP 40--60 mmHg over 0--48 h; O$_2$ flow: $15\rightarrow10$ L/min over 0--48 h.} \\
\revision{9} &
\revision{ECMO support for acute kidney injury} &
\revision{Creatinine: $2\rightarrow150$ mg/dL over 0--96 h; base excess $-30$ to $-20$ mmol/L; HCO$_3^-$ 10--25 mmol/L; SaO$_2$ fixed at $98\%$ over 0--24 h.} \\
\revision{10} &
\revision{ECMO-related complications} &
\revision{Platelet count: $50\rightarrow20\times10^3/\mu$L over 0--96 h.} \\
\bottomrule
\end{tabular}
\end{table}

\revision{The raw parser outputs record requested concepts that are not represented in the interface, such as FiO$_2$, PEEP, lung compliance, plateau pressure, urine output, and ventilator or ECMO settings, under \texttt{unsupported\_facts}. These concepts are not included in the conditioning mask. The raw outputs for scenarios 4 and 8 contain direction labels that conflict with their numerical endpoints, while those for scenarios 4 and 9 include values outside the ranges in the feature dictionary. These outputs require range and direction checks and clinician confirmation or modification before the constraints can be used for generation.}

\section{\revision{\revisiontwo{Metric Definitions for Single-Time-Point Generation}}}
\label{app:single_time_point_metrics}

\revision{%
Formally, let $M$ denote the number of samples in the real dataset and $N$ denote the number of synthetic samples, with $d$ representing the number of features.
Maximum Absolute Mean Difference (MAMD) measures the largest absolute difference between the feature-wise means of real and synthetic data, defined as
\begin{equation}
    \mathrm{MAMD} =
    \max_{j \in \{1,\ldots,d\}}
    \left|
    \hat{\mu}_{\mathrm{syn}}^{(j)}
    -
    \hat{\mu}_{\mathrm{real}}^{(j)}
    \right|.
\end{equation}
Here, $\hat{\mu}_{\mathrm{syn}}^{(j)}$ and $\hat{\mu}_{\mathrm{real}}^{(j)}$ denote the empirical means of feature $j$ in the synthetic and real datasets, respectively:
\begin{align}
    \hat{\mu}_{\mathrm{syn}}^{(j)} &= \frac{1}{N} \sum_{i=1}^{N} x^{(j)}_{i,\mathrm{syn}}, \quad
    \hat{\mu}_{\mathrm{real}}^{(j)} = \frac{1}{M} \sum_{i=1}^{M} x^{(j)}_{i,\mathrm{real}}.
\end{align}
MAMD quantifies the maximum feature-wise mean deviation and is sensitive to feature scales.

The mean-based RMSPE and MAPE metrics are defined as follows:
\begin{align}
    \mathrm{RMSPE} &= 100 \sqrt{\frac{1}{d}\sum_{j=1}^d \Bigg( \frac{\hat{\mu}_{\mathrm{syn}}^{(j)} - \hat{\mu}_{\mathrm{real}}^{(j)}}{\hat{\mu}_{\mathrm{real}}^{(j)}} \Bigg)^2}, \\
    \mathrm{MAPE} &= \frac{100}{d} \sum_{j=1}^d \Bigg| \frac{\hat{\mu}_{\mathrm{syn}}^{(j)} - \hat{\mu}_{\mathrm{real}}^{(j)}}{\hat{\mu}_{\mathrm{real}}^{(j)}} \Bigg|.
\end{align}
}

\end{document}